\documentclass[10pt,journal,compsoc]{IEEEtran}
\ifCLASSOPTIONcompsoc
  \usepackage[nocompress]{cite}
\else
  \usepackage{cite}
\fi
\ifCLASSINFOpdf
\else
\fi

\usepackage{times}
\usepackage{epsfig}
\usepackage{graphicx}
\usepackage{amsmath}
\usepackage{amssymb}
\usepackage{booktabs}
\usepackage{subfigure}
\usepackage{multirow}

\usepackage{amsfonts}
\usepackage{newtxmath}
\usepackage{bbm}

\def\etal{\emph{et al.}}
\newcommand{\argmin}{\operatornamewithlimits{argmin}}
\newcommand{\argmax}{\operatornamewithlimits{argmax}}
\begin{document}
%
\title{Hyperbolic Deep Neural Networks: A Survey}
%
%
%
%

\author{Wei~Peng,~\IEEEmembership{}Tuomas Varanka, Abdelrahman Mostafa, Henglin Shi, 
        Guoying~Zhao*~\IEEEmembership{Senior Member, IEEE}
\IEEEcompsocitemizethanks{\IEEEcompsocthanksitem W. Peng, T. Varanka, A. Mostafa, H. Shi, and G. Zhao are with the Center for Machine Vision and Signal Analysis, University of Oulu, Oulu, Finland.\protect\\
E-mail: firstname.lastname@oulu.fi
}
\thanks{Manuscript received April 19, 2020; revised August 26, 2020.}}

%
%

\markboth{Journal of \LaTeX\ Class Files,~Vol.~14, No.~8, August~2015}%
{Peng \MakeLowercase{\textit{et al.}}: Deep Neural Networks with hyperbolic geometry: A Survey}
%



\IEEEtitleabstractindextext{%
\begin{abstract} 

Recently, there has been a rising surge of momentum for deep representation learning in hyperbolic spaces due to their high capacity of modeling data, like knowledge graphs or synonym hierarchies, possessing hierarchical structure. We refer to the model as hyperbolic deep neural network in this paper. Such a hyperbolic neural architecture can potentially lead to drastically more compact models with much more physical interpretability than its counterpart in the Euclidean space. To stimulate future research, this paper presents a coherent and a comprehensive review of the literature around the neural components in the construction of hyperbolic deep neural networks, as well as the generalization of the leading deep approaches to the hyperbolic space. It also presents current applications around various machine learning tasks on several publicly available datasets, together with insightful observations and identifying open questions and promising future directions.


\end{abstract}

}

\maketitle

\IEEEdisplaynontitleabstractindextext

\IEEEpeerreviewmaketitle

\section{Introduction}\label{sec:introduction}

\IEEEPARstart{L}earning semantically rich representations of data such as text or images is a central point of interest in current machine learning. Recently, deep neural networks~\cite{lecun2015deep} have showcased a great capability in extracting feature representations and have dominated many research fields, such as image classification\cite{he2016deep,deng2009imagenet}, machine translation tasks~\cite{vaswani2017attention}, and playing video games~\cite{mnih2013playing}. Theoretically, deep neural networks with millions of parameters have great potential to fit any complex functions, especially when they are equipped with many advanced optimization libraries~\cite{abadi2016tensorflow,paszke2017automatic,paszke2019pytorch}. At the core of deep neural networks lies the expectation to find the optimal representations (task-related), of which the underlying manifold assumption is that the intrinsic dimensionality of the input data is much lower compared to the input feature space dimension. In most of the current deep learning applications, the representation learning is conducted in the Euclidean space, which is reasonable since the Euclidean space is the natural generalization of our intuition-friendly, visual three-dimensional space. However, recent research has shown that many types of complex data exhibit a highly non-Euclidean latent anatomy~\cite{bronstein2017geometric}. Besides, it appears in several applications that the dissimilarity measures constructed by experts tend to have non-Euclidean behavior~\cite{duin2010non}. In such cases, the Euclidean space does not provide the most powerful or meaningful geometrical representations. Works like~\cite{lee2013smooth} even suggest that data representations in most machine learning applications lie on a smooth manifold, and the smooth manifold is Riemannian, which is equipped with a positive definite metric on each tangent space, i.e., every non-vanishing tangent vector has a positive squared norm. Due to the positive definiteness of the metric, the negative of the (Riemannian) gradient is a descent direction that can be exploited to iteratively minimize some objective functions. Therefore current research is increasingly attracted by the idea of building neural networks in a Riemannian space, such as the hyperbolic space~\cite{gromov1987hyperbolic,WikiHyperbolic}, which is a Riemannian manifold with constant negative (sectional) curvature, and it is analogous to a high-dimensional sphere with constant positive curvature. 

In many domains, data is with a tree-like structure or can be represented hierarchically~\cite{zhu2016generative}. For instance, social networks, human skeletons, sentences in natural language, and evolutionary relationships between biological entities in phylogenetics. As also mentioned by~\cite{shimizu2021hyperbolic}, a wide variety of real-world data encompasses some types of latent hierarchical structures~\cite{newman2005power,lin2017critical,katayama2015indexing}. Besides, from the perspective of cognitive science, it is widely accepted that human beings use hierarchy to organise object categories ~\cite{collins1969retrieval,keil2013semantic,roy2007learning}. As a result, there is always a passion to model the data hierarchically. Explicitly incorporating hierarchical structure in probabilistic models has unsurprisingly been a long-running research topic. Earlier work in this direction tended towards using trees as data structures to represent hierarchies. Recently, hyperbolic spaces have been proposed as an alternative continuous approach to learn hierarchical representations from textual and graph-structured data. The negative-curvature of the hyperbolic space results in very different geometric properties, which makes it widely employed in many settings. In the hyperbolic space, circle circumference ($2 \text{~sinh~} r$) and disc area ($2\pi(\text{~cosh~} r-1)$) grow
exponentially with radius $r$, as opposed to the Euclidean space where they only grow linearly and quadratically. The exponential growth of the Poincar\'e surface area with respect to its radius is analogous to the exponential growth of the number of leaves in a tree with respect to its depth, rather than polynomially as in the Euclidean case~\cite{anderson2006hyperbolic,ganea2018hyperbolic}. On the contrary, Euclidean space, as shown in Bourgain’s theorem~\cite{linial1995geometry}, is unable to obtain comparably \textbf{low distortion} for tree data, even when using an unbounded number of dimensions. 
Furthermore, the hyperbolic spaces are smooth, enabling the use of deep learning approaches which rely on differentiability. Therefore, hyperbolic spaces have recently gained momentum in the context of deep neural networks to model embedded data into the space that exhibits certain desirable geometric characteristics. To summarize, there are several potential advantages of utilizing hyperbolic deep neural networks to represent data:

\begin{itemize}

\item  A better generalization capability of the model, with less overfitting, computational complexity, and requirement of training data. 
\item Reduction in the number of model parameters and embedding dimensions. 
\item A low distortion embedding, which preserves the local and geometric information.
\item A better model understanding and interpretation.  
\end{itemize}

Hyperbolic deep neural networks have a great potential in both academia and industry. For instance, in academia, the negatively curved spaces have already gained much attention for embedding and representing relationships between objects that are organized hierarchically~\cite{nickel2017poincare,ganea2018hyperbolic}. In industry, there are already recommender systems based on hyperbolic models, which are scaled to millions of users~\cite{chamberlain2019scalable}. Also, we can find work for learning drug hierarchy~\cite{yu2020HierarchicalDrug}. Although constructing neural networks on hyperbolic space gained great attention,  as far as we know, there is currently no survey paper in this field. This article makes the first attempt and aims to provide a comprehensive review of the literature around hyperbolic deep neural networks for machine learning tasks. Our goals are to 1) provide concise context and explanation to enable the reader to become familiar with the basics of hyperbolic geometry, 2) review the current literature related to algorithms and applications, and 3) identify open questions and promising future directions. We hope this article could be a tutorial for this topic, as well as a formal theoretical support for future research.

The article is organized as follows. In Section~\ref{Sec2}, we introduce the fundamental concepts about hyperbolic geometry, making the paper self-contained. 
Section~\ref{Sec4} introduces the generalization of important neural network components from the Euclidean space to the hyperbolic space. We then review the constructions for hyperbolic deep neural networks, including building networks on two commonly used hyperbolic models, Lorentz Model and Poincar\'e Model in Section~\ref{Sec5}. In Section~\ref{Sec6}, we describe applications for testing hyperbolic deep neural networks and discuss the performance of different approaches under different settings. Finally, in Section~\ref{Sec7} we identify open problems and possible future research directions.


\section{Hyperbolic Geometry}\label{Sec2}
In this section, we briefly introduce the background of hyperbolic geometry. Then, we discuss the mathematical preliminaries and notations of hyperbolic geometry. Finally, we go through the commonly used isometric models in the hyperbolic space. 

\subsection{Background}
Hyperbolic Geometry is a non-Euclidean geometry, also called the Lobachevsky-Bolyai-Gauss geometry, having constant sectional curvature of $-1$~\footnote{Note that, for a clear description,  we fix the curvature of models to -1. This can be easily generalized to a hyperbolic space with other (negative) curvature.}. This geometry satisfies all of Euclid's Postulates except the parallel postulate. Non-Euclidean geometry arises by either relaxing the metric requirement or by replacing the parallel postulate with an alternative. Before we go any further, we will talk about the postulates first. In fact, the Euclidean space is also constructed upon some postulates, which are the well-known Euclid's postulates. Postulates in geometry are very similar to axioms, self-evident truths, and beliefs in logic, political philosophy and personal decision-making. The five postulates of Euclidean Geometry define the basic rules governing the creation and extension of geometric figures with a ruler and compass. They are as follows:

\begin{itemize}
    \item A straight line segment can be drawn joining any two points.
    \item Any straight line segment may be extended to any finite length.
    \item A circle may be described with any given point as its center and any distance as its radius.
    \item All right angles are congruent.
    \item  If two lines are drawn which intersect a third in such a way that the sum of the inner angles on one side is less than two right angles, then the two lines inevitably must intersect each other on that side if extended far enough. This postulate is equivalent to what is known as the parallel postulate.
\end{itemize}

\begin{figure*}
    \centering
    \includegraphics[width=0.9\textwidth]{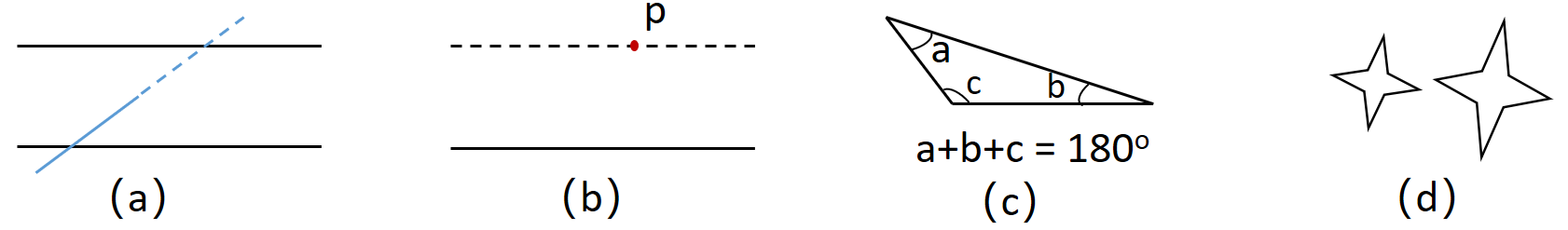}
    \caption{The equivalent statements for the fifth postulate.(a) If a straight line intersects one of the two parallels, it will intersect the other also. (b) There is one and only one line that passes through any given point and is parallel to the given line. (c) There is a triangle in which the sum of the three angles is $180^o$. (d) Given any figure, there exists a figure, of any size.}
    \label{fig:5postlicate}
\end{figure*}

To better help understand the last postulate Fig~\ref{fig:5postlicate} shows equivalent statements pictorially.
Here we take a look at the parallel postulate, which is shown in Fig.~\ref{fig:5postlicate} (b) and its definition is that given any straight line and a point not on it, ``there exists one and only one straight line which passes through that point and never intersects the first line, no matter how far they are extended''.  This statement is equivalent to the fifth postulate and it was the source of much annoyance, and for at least a thousand years, geometers were troubled by the disparate complexity of the fifth postulate.  As a result, many mathematicians over the centuries have tried to prove the results of the Elements without using the Parallel Postulate, but to no avail. However, in the past two centuries, assorted non-Euclidean geometries have been derived based on using the first four Euclidean postulates together with various negations of the fifth. To obtain a non-Euclidean geometry, the parallel postulate (or its equivalent) is replaced by its negation. As stated above, the parallel postulate describes the type of geometry now known as parabolic geometry. If, however, the phrase ``exists one and only one straight line which passes'' is replaced by ``exists no line which passes,'' or ``exists at least two lines which pass,'' the postulate describes equally valid (though less intuitive) types of geometries known as elliptic and hyperbolic geometries, respectively.

The model for hyperbolic geometry was answered by Eugenio Beltrami, in 1868, who first showed that a surface called the pseudo-sphere has the appropriate curvature to model a portion of the hyperbolic space and in a second paper in the same year, defined the Klein model, which models the entirety of hyperbolic space, and used this to show that Euclidean geometry and hyperbolic geometry are equiconsistent (as consistent as each other) so that hyperbolic geometry is logically consistent if and only if Euclidean geometry is~\cite{WikiHyperbolic}.

\subsection{Mathematical preliminaries}

Here, we provide preliminaries notations to keep this paper self-contained.

    \textbf{Manifold}: A manifold $\mathcal{M}$ of dimension $n$ is a topological space of which each point's neighborhood can be locally approximated by the Euclidean space $\mathbb{R}^n$. For instance, the earth can be modeled by a sphere, while its local place looks like a flat area which can be approximated by $\mathbb{R}^2$. The notion of manifold is a generalization of the notion of surface.
    
    \textbf{Tangent space}: For each point $x \in \mathcal{M}$, the tangent space $\mathcal{T}_x\mathcal{M}$ of $\mathcal{M}$ at $x$ is defined as a $n$-dimensional vector-space approximating $\mathcal{M}$ around $x$ at a first order. It is isomorphic to $\mathbb{R}^n$. It can be defined as the set of vectors $v$ that can be obtained as $v := c_0(0)$, where $c: (-\epsilon, \epsilon) \rightarrow \mathcal{M} $ is a smooth path in  $\mathcal{M}$ such that $c(0) = x$.
    
    \textbf{Riemannian metric}:  The metric tensor gives a local notion of angle, length of curves, surface area, and volume. For a manifold $\mathcal{M}$, a Riemannian metric $\mathfrak{g}(x)$ is a smooth collection of inner products on the associated tangent space: $<\cdot,\cdot>_x: \mathcal{T}_x\mathcal{M} \times \mathcal{T}_x\mathcal{M}$.
    
    \textbf{Riemannian manifold}: A Riemannian manifold~\cite{carmo1992riemannian} is then defined as a manifold equipped with a group of Riemannian metrics $\mathfrak{g}$, which is formulated as a tuple $(\mathcal{M}$, $\mathfrak{g})$~\cite{gallot1990riemannian}.
    
    \textbf{Geodesics}: Geodesics is the the generalization of a straight line in the Euclidean space. It is the constant speed curve giving the shortest (straightest) path between pairs of points.
    
    \textbf{Isomorphism}: An isomorphism is a structure-preserving mapping between two structures of the same type that can be reversed by an inverse mapping.
    
    \textbf{Homeomorphism}: A homeomorphism is a continuous one-to-one and onto mapping that preserves topological properties. Homeomorphisms are isomorphisms of topological spaces. 

    \textbf{Conformality}: Conformality is one important property of the Riemannian geometry. A metric $\hat{\mathfrak{g}}$ on a manifold $\mathcal{M}$ is conformal to the $\mathfrak{g}$ if it defines the same angles. This is equivalent to the existence of a smooth function  $\lambda:  \mathcal{M} \rightarrow (0, \infty)$ such that $\hat{\mathfrak{g}}_x = \lambda_x^2 \mathfrak{g}_x$. Formally, $\forall x \in \mathcal{M}$ and $u,v \in \mathcal{T}_x\mathcal{M}\backslash \{0\}$:
    \begin{equation}
        \frac{\hat{\mathfrak{g}}_x(u,v)}{\sqrt{\hat{\mathfrak{g}}_x(u,u)}\sqrt{\hat{v}_x(v,v)}} = \frac{{\mathfrak{g}}_x(u,v)}{\sqrt{{\mathfrak{g}}_x(u,u)}\sqrt{{\mathfrak{g}}_x(v,v)}}.
    \end{equation}

    \textbf{Exponential map}: The exponential map takes a vector $v \in \mathcal{T}_x\mathcal{M}$ of a point $x \in \mathcal{M}$ to a point on the manifold $\mathcal{M}$, i.e., $\text{Exp}_x:  \mathcal{T}_x\mathcal{M} \rightarrow  \mathcal{M}$ by moving a unit length along the geodesic uniquely defined by $\gamma(0) = x$ with direction $\gamma^{'} (0) = v$. Different manifolds have their own way to define the exponential maps. Generally, this is very useful when computing the gradient, which provides update that the parameter moves along the geodesic emanating from the current parameter position.
    
    \textbf{Logarithmic map}: As the inverse of the aforementioned exponential map, the logarithmic map projects a point $z \in \mathcal{M}$ on the manifold back to the tangent space of another point $x \in \mathcal{M}$, which is $\text{Log}_x:  \mathcal{M} \rightarrow  \mathcal{T}_x\mathcal{M}$. Like the exponential map, there are also different logarithmic maps for different manifolds.
    
    \textbf{Parallel Transport}: Parallel Transport defines a way of transporting the local geometry along smooth curves that preserves the metric tensors. In particular, the parallel transport $\mathcal{PT}_{u \rightarrow v}$ from vector $u \in \mathcal{M}$ to $v \in \mathcal{M}$ is defined as a map from tangent space of $u$, $\mathcal{T}_u \mathcal{M}$ to $\mathcal{T}_v \mathcal{M}$ that carries a vector in $\mathcal{T}_u \mathcal{M}$ along the geodesic from $u$ to $v$.

    \textbf{Curvature}: There are multiple notions of curvature, including scalar curvature, sectional curvature, and Ricci curvature, with varying granularity. Note that curvature is inherently a notion of two-dimensional surfaces, while the sectional curvature fully captures the most general notion of curvature (the Riemannian curvature tensor). And The Ricci curvature of a tangent vector is the average of the sectional curvature over all planes $U$ containing $v$. Scalar curvature is a single value associated with a point $p \in \mathcal{M}$ and intuitively relates to the area of geodesic balls. Negative curvature means that volumes grow faster than in the Euclidean space, and the positive curvature is the opposite of volumes growing slower. The sectional curvature varies over all “sheets” passing through $p$ and intuitively, it measures how far apart two geodesics emanating from $p$ diverge. In positively curved spaces like the sphere, they diverge more slowly than in flat Euclidean space. The Ricci curvature, geometrically measures how much the volume of a small cone. Positive curvature implies smaller volume, and negative implies larger. 
    
    
    \textbf{Gromov $\delta$-hyperbolicity}: Gromov $\delta$-hyperbolicity~\cite{adcock2013tree} is used to evaluate the hyperbolicity of a dataset/space. Normally, it is defined under four-point condition, say points $a,b,c,v$.  A metric space $(X, d)$ is $\delta$-hyperbolic if there exists a $\delta > 0$ such that these four points in X: $<a,b>_v \geq \min \{<a,c>_v,<b,c>_v\}- \delta$, where the $<,>_v$ with respect to a third point v is the Gromov product~\cite{gromov1987hyperbolic} of points $a,b$ and it is defined as $<a,b>_v = \frac{1}{2}(d(a,v)+d(b,v)-d(a,b))$. For instance, Euclidean space $\mathbb{R}^n$ is not $\delta$-hyperbolic, Poincar\'e disc ($\mathbb{B}^2$) is (log3)-hyperbolic.
    
    \textbf{Average Distortion}: is a standard measurement of various fidelity measures. It is commonly used in feature embedding. Let $U, V$ be two metric spaces equipped with distances $d_U$ and $d_V$, and the embedding is the function $f: U \rightarrow V$. Then the distortion $\mathbb{D}$ is defined as $\mathbb{D} = \frac{|d_U(a,b)- d_V(f(a),f(b))|}{d_U(a,b)}$, for a pair of points $a,b$ on the space of $U$.

\subsection{Five Isometric Models in the Hyperbolic Space}\label{Sec3}

Hyperbolic space is a homogeneous space with a constant negative curvature. It is a smooth Riemannian manifold and as such locally Euclidean space. The hyperbolic space can be modelled using five isometric models~\cite{beltrami1868teoria,cannon1997hyperbolic}, which are the Lorentz (hyperboloid) model, the Poincar\'e ball model, Poincar\'e half space model, the Klein model, and the hemishpere model. They are embedded sub-manifolds of ambient real vector spaces. In the following parts, we will detail these models one-by-one. Note that we describe the model by fixing the radius of the model to 1 for clarity, without loss of generality.

\begin{figure*}
    \centering
    \includegraphics[width = 0.8 \textwidth]{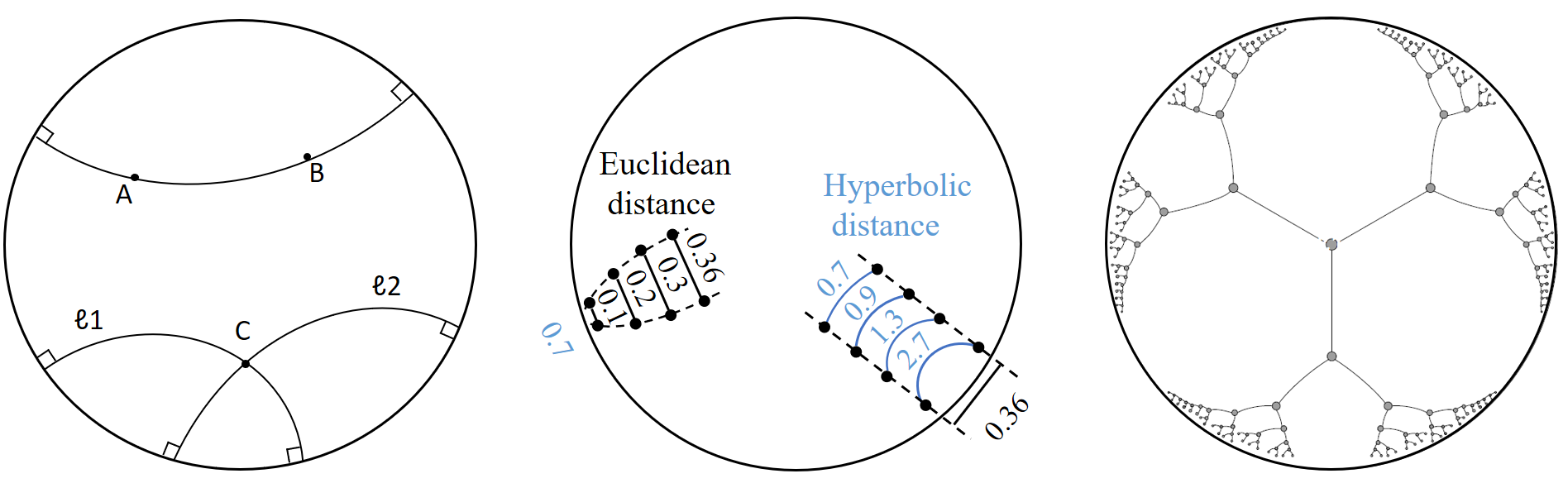}
    \caption{Illustration of Poincar\'e Disk (2D Poincar\'e model), Leftmost: the 'straight lines' in Poincar\'e model. Going through a point c outside of line AB, there are more than one lines paralleling to line AB. Middle: the distances comparison between Euclidean space (in black) and hyperbolic space (in blue). Rightmost: an example of modeling a tree using hyperbolic model.}
    \label{fig:poincare}
\end{figure*}

\subsubsection{Lorentz Model}

The Lorentz model $\mathbb{L}^n$ of an $n$ dimensional hyperbolic space is a manifold embedded in the $n + 1$ dimensional Minkowski space. The Lorentz model is defined as the upper sheet of a two-sheeted n-dimensional hyperbola with the metric $\mathfrak{g}^{L}$, which is

\begin{equation}
    \mathbb{L}^n = \{ x=(x_0,...,x_{n}) \in \mathbb{R}^{n+1}: <x,x>_{\mathbb{L}} = -1, x_0 > 0\},
\end{equation}
where the $<,>_{\mathbb{L}}$ represents the Lorentzian inner product, which is defined as

\begin{equation}
 <x,y>_{\mathbb{L}} = x^T\mathfrak{g}^{L}y = -x_0y_0 + \sum_{i=1}^{n}x_iy_i, x~ \text{and}~ y \in \mathbb{R}^{n+1},
\end{equation} where $\mathfrak{g}^L$ is a diagonal matrix with entries of 1s, except for the first element being -1. For any $x \in \mathbb{L}^n$, we can get that $x_0 = \sqrt{1+ \sum_{i=1}^{n+1} x_i^2}$. The distance in the Lorentz Model is defined as
\begin{equation}
    d(x,y) = \text{~arcosh~}(-<x,y>_{\mathbb{L}}).
\end{equation}

The main advantage of this parameterization model is that it provides an efficient space for Riemannian optimization. An additional advantage is that its distance function avoids numerical instability, when compared to Poincar\'e model, in which the instability arises from the fraction.

\subsubsection{Poincar\'e Model}\label{sec:pball}

The Poincar\'e model is given by projecting each point of $\mathbb{L}^n$ onto the hyperplane $x_0 = 0$, using the rays emanating from (-1, 0,..., 0). The Poincar\'e model $\mathbb{B}$ is a manifold equipped with a Riemannian metric $\mathfrak{g}^{B}$. This metric is conformal to the Euclidean metric $\mathfrak{g}^E$ with the conformal factor $\lambda_x = \frac{2}{1-||x||^{2}}$, and $\mathfrak{g}^B = \lambda_x^2 \mathfrak{g}^E$. Formally, an $n$ dimensional Poincar\'e unit ball (manifold) is defined as
\begin{equation}
    \mathbb{B}^n = \{ x\in \mathbb{R}^{n}: ||x|| < 1\},
\end{equation}
where $||\cdot||$ denotes the Euclidean norm. Formally, the distance between $x,y \in \mathcal{M}$ is defined as:
\begin{equation}
    d(x,y) = \text{~arcosh~}\left(1+2\frac{||x-y||^2}{(1-||x||^2)(1-||y||^2)}\right).
\end{equation}

As illustrated in Fig.~\ref{fig:poincare}, we demonstrate the geometry of this manifold in the two-dimensional model, Poincar\'e disk. The leftmost shows how to generalize a straight line in this model. The middle compares the hyperbolic distance between two points (curves in blue) to that in the Euclidean space(lines in black). Like in the left line group, given a constant hyperbolic distance 0.7, the corresponding Euclidean distances decrease dramatically when the points are closed to the unit boundary. This exponentially growing distance fit very well with the depth increasing in a tree, as such the Poincar\'e disk is very suitable for modeling a tree, as shown in the rightmost of Fig.~\ref{fig:poincare}. 

\subsubsection{Poincar\'e half plane Model}
The closely related Poincar\'e half-plane model in hyperbolic space is a Riemannian manifold $(\mathbb{H}^n, \mathfrak{g}^H)$, where 
\begin{equation}
    \mathbb{H}^n = \{ x\in \mathbb{R}^{n}: x_n >0 \}
\end{equation}
is the upper half space of an $n$-dimensional Euclidean space. And the metric $\mathfrak{g}^H$ is given by scaling the euclidean metric  $\mathfrak{g}^H = \frac{\mathfrak{g}^E}{x_n^2}$. The model $\mathbb{H}^n$ can be obtained by taking the inverse of Poincar\'e model, $\mathbb{B}^n$, with respect to a circle that has a radius twice that of $\mathbb{B}^n$.
The distance is 
\begin{equation}
    d(x,y) = \text{~arcosh~}\left(1+ \frac{||x-y||^2}{2x_n y_n} \right).
\end{equation}

\subsubsection{Klein Model}

Klein model is also known as the Beltrami–Klein model, named after the Italian mathematician Eugenio Beltrami and German mathematician Felix Klein. The Klein model of hyperbolic space is a subset of $\mathbb{R}^n$. It can also be treated as an projection of Lorentz model. Instead of projecting onto the hyperplane $x_0 = 0$, the Klein ball can be obtained by mapping $x \in \mathbb{L}^{n+1}$ to the hyperplane $x_0 = 1$, using rays emanating from the origin. Formally, the Klein model is defined as

\begin{equation}
    \mathbb{K}^n = \{ x\in \mathbb{R}^{n}: ||x|| < 1 \},
\end{equation}

The distance is 
\begin{equation}
    d(x,y) = \text{~arcosh~}(1+ \frac{(y_0-x_0)^2+(y_1-x_1)^2}{2x_1y_1}).
\end{equation}

\subsubsection{Hemisphere model}

The hemisphere model is also called Jemisphere model, which is not as common as the previous four models. Instead, this model is employed as a useful tool for visualising transformations between other models. Hemisphere model is defined as 
\begin{equation}
    \mathbb{J}^n = \{ x=(x_0,...,x_{n}) \in \mathbb{R}^{n+1}: ||x|| = 1, x_0 > 0\},
\end{equation}

\subsubsection{Isometrically Equivalent Models}\label{sec:isom}
\begin{figure}
    \centering
    \includegraphics[width=0.4\textwidth]{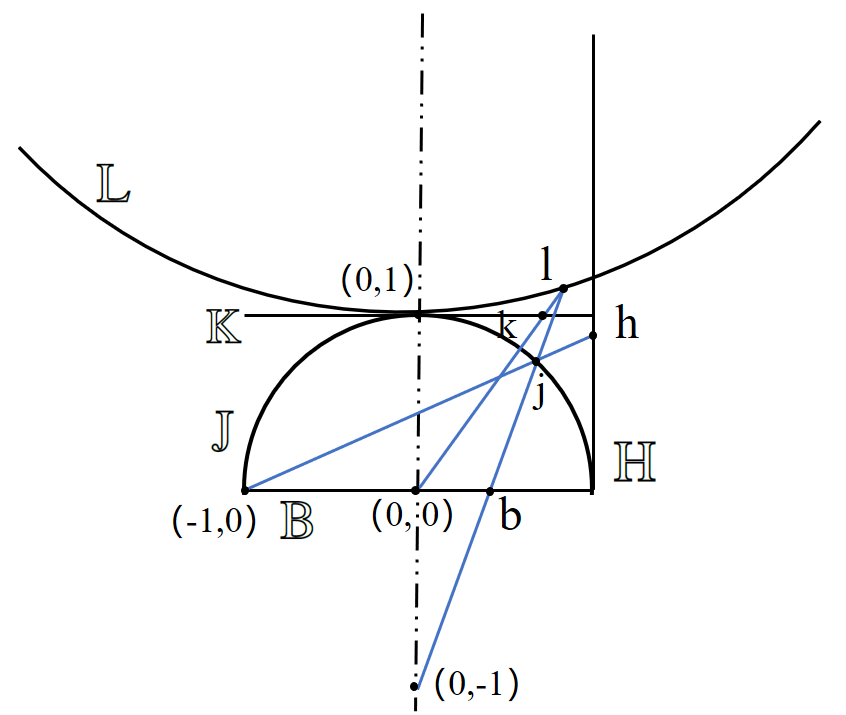}
    \caption{Illustration of the relationship in the five hyperbolic models. Here, the five models are represneted in a two dimensional space. The points $h\in \mathbb{H}$, $b \in \mathbb{B}$, $j \in \mathbb{J}$, $k \in \mathbb{K}$, and $l \in \mathbb{L}$ can be thought of as the same point in the hyperbolic space.}
    \label{fig:five_model_r}
\end{figure}

In fact, these five models are equivalent models of the hyperbolic space. There are closed-form expressions for mapping between these hyperbolic models. As illustrated in Fig.~\ref{fig:five_model_r}, we display their model in a 2-dimensional space and demonstrate their relationship. 

Here, we describe the isometries among them. Lorentz model can be transferred to  Poincar\'e model by this mapping function: 
\begin{equation}
    x=(x_0,...,x_{n})\in \mathbb{L}^n \Leftrightarrow \left(\frac{x_1}{1+x_0}, \ldots, \frac{x_{n}}{1+x_0}\right)\in \mathbb{B}^n.
\end{equation}

Likewise, we can further construct a mapping from Poincar\'e model to Poincar\'e half plane Model by inversion on a circle centered at $(1,0,...,0)$, the resulting correspondence between $\mathbb{B}^n$ and $\mathbb{H}^n$ is given by

\begin{equation}
    x=(x_0,...,x_{n-1})\in \mathbb{B}^n \Leftrightarrow \left(\frac{1-||x||^2, 2x_1,...,2x_{n-1}}{1+2x_0+||x||^2}\right)\in \mathbb{H}^n.
\end{equation}


As mentioned before, Klein model is also a projection of Lorentz model, the transform relationship is

\begin{equation}
    x=(x_0,...,x_{n})\in \mathbb{L}^n \Leftrightarrow \left(\frac{x_1}{x_0}, ..., \frac{x_{n}}{x_0}\right)\in \mathbb{K}^n.
\end{equation}

As mentioned in~\cite{cannon1997hyperbolic}, the points on the hemisphere model can be projected to 

\begin{equation}
    x=(x_0,...,x_{n})\in \mathbb{L}^n \Leftrightarrow \left(\frac{x_0}{x_{n+1}}, ..., \frac{x_{n}}{x_{n+1}}, \frac{1}{x_{n+1}}\right)\in \mathbb{J}^{n+1}.
\end{equation}


\section{Generalizing Euclidean Operations to the Hyperbolic Space}\label{Sec4}

The hyperbolic space is endowed with various geometric properties~\cite{nickel2017poincare,ganea2018hyperbolic}, such that it has the potential to alleviate some of the current machine learning problems for certain types of data. Furthermore, currently, optimization methods essential for any machine learning pipeline have recently made great advances. Therefore, generalizing fundamental operations from Euclidean space to hyperbolic space is one well-motivated approach to designing better machine learning architectures. Although the use of hyperbolic embeddings (first proposed by Kleinberg~\etal ~\cite{kleinberg2007geographic}) in machine learning was introduced already early in 2007, only recently have the methods been extended to deep neural networks. Constructing deep neural networks in the hyperbolic space is not as easy as it is on the Euclidean space. One of the most crucial reasons that hampers the development of hyperbolic counterparts of deep neural architectures is the difficulty of developing operations in the hyperbolic space required for neural networks. To summarize, there are several challenges making this a nontrivial issue:

\begin{itemize}

\item Implementations of modeling, learning, and optimization on Riemannian manifolds are not as efficient as they are in the Euclidean space.
\item It is challenging to obtain closed form expressions of the most relevant Riemannian geometric tools, such as geodesics, exponential maps, or distance functions, since these geometric elements can easily lose their appealing closed form expressions.
\item The adoption of neural networks and deep learning in these non-Euclidean settings has been rather limited until very recently.

\end{itemize}

One of the main reasons is that it is being the nontrivial or impossible principled generalizations of basic operations, e.g., vector addition, matrix-vector multiplication. Work~\cite{ganea2018hyperbolic} provided a pioneer study of how classical Euclidean deep learning tools can be generalized in a principled manner to hyperbolic space. Fueled by this, many current works generalize various deep learning operations as it is the key step towards to hyperbolic deep neural networks. In this section, we will review the research literature which is trying to generalize operations, e.g., basic addition, mean and neural network layers, to the hyperbolic space.

One easy and straightforward way to generalize all these neural operations is transferring data in the hyperbolic space to a tangent space, where we build all operations just like we do in the Euclidean space, as tangent space keeps local Euclidean properties. However, as noticed in some works~\cite{lou2020differentiating,chami2019hyperbolic}, the approximation in the tangent space can have a negative impact on the learning process. Thus, more advanced approaches, like directly building neural operations using hyperbolic geometry, are very much expected. We will detail these methods in the corresponding sections.

\subsection{Basic Arithmetic Operations}

Basic mathematical operations, like addition and multiplication, are fundamental components of neural networks. They are everywhere in the neural network components, like convolutional filters, fully connected layers, and activation functions.

As mentioned before, one simple way to perform these computations is to approximate them by employing the tangent space.

Another good choice is the {Gyrovector space}~\cite{ungar2001hyperbolic}, which is a generalization of Euclidean vector spaces to models of hyperbolic space based on M\"{o}bius transformations. Specifically, for a model $\mathbb{B}:=\{x \in \mathbb{R}^n, ||x|| < 1\}$, the gyrovector space provides a non-associatitative algebraic formulation for studying hyperbolic geometry, in analogy to the way vector spaces are used in Euclidean geometry. 

In the Gyrovector space, the \textbf{M\"{o}bius addition} $\oplus$ for x and y in model $\mathbb{B}$ is defined as
\begin{equation}
    x\oplus y = \frac{(1+2\left \langle x,y  \right \rangle+||y||^2)x+(1-||x||^2)y}{1+2\left \langle x,y  \right \rangle +||x||^2||y||^2}.
\end{equation}

This is a generalization of the addition in Euclidean space. And $ x\oplus y $ will recover to $x+y$ when the curvature goes to zero. In addition, the M\"{o}bius subtraction is simply defined as: $x \ominus y = x\oplus (-y) $.

Then the \textbf{M\"{o}bius scalar multiplication} $\otimes$ is defined as

\begin{equation}
  r \otimes x = \left\{
   \begin{matrix}
 & \text{~tanh~}(r \text{~artanh~} (||x||)\frac{x}{||x||}, &~x \in \mathbb{B}^n \\ 
 & 0, &~x=0,
\end{matrix}\right.
\end{equation}
where $r$ is a scalar factor. In fact, all above-mentioned operations can also be conducted in the tangent space by using the exponential and logarithmic maps. As provided by~\cite{gulcehre2018hyperbolic}, the \textbf{M\"{o}bius scalar multiplication} can be obtained by projecting x in the tangent space at 0, multiplying this projection by the scalar r in the tangent space. Then projecting it back on the manifold with the exponential map, which means 
\begin{equation}
  r \otimes x = \text{Exp}_0(r \text{Log}_0(x)).
\end{equation}

With the similar method, the authors derived the \textbf{M\"{o}bius vector multiplication} $M^{\otimes}(x)$ between the matrix $M$ and input $x$, which is defined as
\begin{equation}
    M^{\otimes}(x) = \text{~tanh~}\left( \frac{||Mx||}{||x||}\text{~actanh~}(||x||)\right) \frac{Mx}{||Mx||}
\end{equation}
Based on the M\"{o}bius tranformations, the authors~\cite{ganea2018hyperbolic} also derived a closed from expression of {M\"{o}bius exponential and logarithmic maps} for Pincar\'e model. For a vector $v \in \mathcal{T}_x\mathcal{M}$ in the tangent space, the exponential map is defined as 
\begin{equation}
    \text{Exp}_x(v) = x \oplus (\text{~tanh~}(\frac{\lambda_x||v||}{2})\frac{v}{||v||}),
\end{equation}
and as the inverse operation of the exponential map, for a point $y \in \mathbb{B}$ on the manifold, the logarithmic map is defined as 
\begin{equation}
    \text{Log}_x(y) = \frac{2}{\lambda_x}\text{~artanh~}(||-x\oplus y||)\frac{-x\oplus y}{||-x\oplus y||}.
\end{equation}

In which the $\lambda_x$ is the conformal factor, as mentioned in Sec.~\ref{sec:pball}.

\subsection{Mean in the Hyperbolic Space}\label{sec:mean}
The simple but valuable mean computation is one of the most fundamental operation in machine learning approaches. For instance, the average pooling in deep learning, the statistics of a data (feature) distribution, and information aggregation in graph convolutional networks. However, unlike in the Euclidean space, the mean computation cannot be conducted simply by averaging the inputs, which may lead to a result out of the manifold. Basically, the primary baselines to generalize the mean to hyperbolic space are tangent space aggregation~\cite{chami2019hyperbolic}, Einstain midpoint method~\cite{gulcehre2018hyperbolic}, and the Fr\'echet mean method~\cite{lou2020differentiating}. We will detail these methods in the following part.

\textbf{Tangential aggregations} is one of the most straightforward ways to compute the mean in hyperbolic space. It was proposed by work~\cite{chami2019hyperbolic}, in which this mean computation method is introduced to perform information aggregation for hyperbolic graph covolutional network (GCN). Generally, the mean aggregation in Euclidean GCN is defined as a weighted average $\mu$ on the involved neighbor nodes, $\mathcal{N}(i)$, of node i, which is 
\begin{equation}
    \mu = \sum_{j \in \mathcal{N}(i)} w_j x_j.
\end{equation}
However, directly computing the weighted average in the hyperbolic space is not able to ensure the resulting average would still be on the manifold. Thus, work~\cite{chami2019hyperbolic} turns to the tangent space by using projection functions, and an attention based aggregation is proposed to compute the aggregated information. Specifically, given the corresponding hyperbolic feature representation, one can compute the attention weights $w_{ij}$ first, then the mean (aggregated information) $\mu$ is
\begin{equation}
    \mu = \text{Exp}_x( \sum_{j \in \mathcal{N}(i)} w_{ij} \text{Log}_x( x_j)).
\end{equation}

Instead of approximating the mean in tangent space, work~\cite{gulcehre2018hyperbolic} proposes to compute it with \textbf{Einstein midpoint}. Einstein midpoint is an extension of the mean operation to hyperbolic spaces, which has the most concise form with the Klein coordinates. The Einstein midpoint is defined as
\begin{equation}
    \mu = \frac{\sum_{i=1}^{N}\gamma_i x^{(i)}}{\sum_{i=1}^{N}\gamma_i},
\end{equation}
in which the $x^{(i)}$ is the i-th sample represented with coordinates in Klein model. The $\gamma_i = \frac{1}{||x^{(i)}||^2}$ are the Lorentz factors. One can easily execute midpoint computations by simply projecting to and from the Klein model to various models of hyperbolic space since all of them are isomorphic.

\begin{figure}
    \centering
    \includegraphics[width = 0.4 \textwidth]{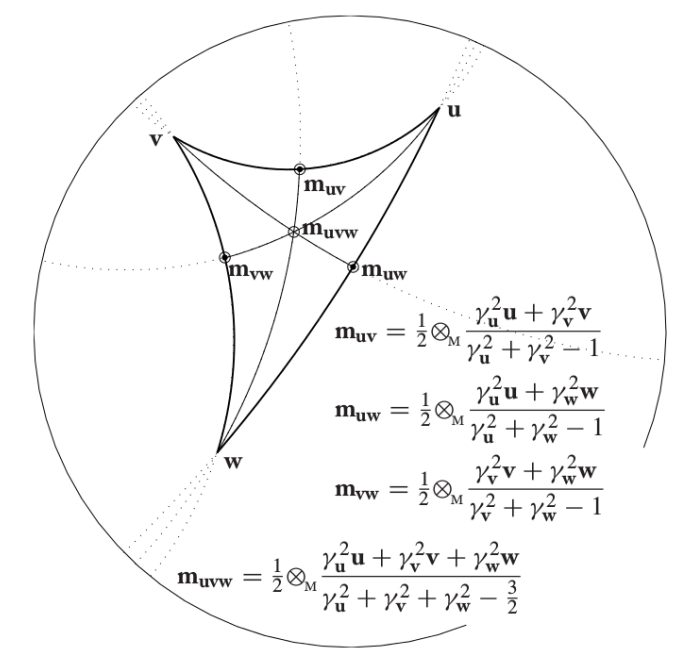}
    \caption{Illustration of the M\"{o}bius gyromidpoint. Here we show it in the Poincar\'e model, which is defined by work~\cite{ungar2008gyrovector}.}
    \label{fig:gmidpoint}
\end{figure}

In fact, there is also a closed form expression for Poincar\'e model to compute the average (midpoint) in the gyrovector spaces. Work~\cite{ungar2008gyrovector} defines a gyromidpoint, as illustrated in Fig.~\ref{fig:gmidpoint}. The definition is 

\begin{equation}
    m(x^{(1)},...,x^{(N)};\alpha) = \frac{1}{2} \oplus \left( \sum_{i=1}^{N} \frac{\alpha_i \gamma_i}{\sum_{j=1}^{N} \alpha_j (\gamma_j -1)} x^{(i)}\right)
\end{equation}
with $\alpha = (\alpha_1,..., \alpha_N)$ as the weights for each sample $x^{(i)}$ and $\gamma_i = \frac{2}{||x^{(i)}||^2}$.

Work~\cite{gulcehre2018hyperbolic} developed a new method to perform the information aggregation in the hyperbolic space based on the Einstein midpoint.

\textbf{Fr\'echet mean}
Recently, Luo~\etal~proposed a new method to compute the mean on the hyperbolic space, which derives a closed-form gradient expressions for the Fr\'echet mean ~\cite{Frechetmean1948} on Riemannian manifolds~\cite{lou2020differentiating}. Based on this, they presented two neural operations in hyperbolic space and achieved superior performance under the given settings. Firstly, they provided a new way to conduct the information aggregation for graph convolutional network in the hyperbolic space. Secondly, they generalized the standard Euclidean batch normalization. We will detail them in corresponding sections. Here, we introduce how they provided a differentiable Fr\'echet mean.

In fact, there are considerable years for generalizing Euclidean mean in non-Euclidean geometries~\cite{frechet1948elements}, using Fr\'echet mean. However, the Fr\'echet mean does not have a closed-form solution, and its computation involves an argmin operation that cannot be easily differentiated. Besides, as mentioned by work~\cite{lou2020differentiating}, computing the Fr\'echet mean relies on some iterative solver~\cite{lou2020differentiating}, which is computationally
inefficient, numerical instability and thus not friendly to deep neural networks. Thus, the authors derived a optimization objective for mean (and variance) computation in hyperbolic space, which is 
\begin{equation}\label{eq:mean}
    \mu_{fr} = \argmin_{\mu \in \mathcal{M}} \frac{1}{t}\sum_{l=1}^{t} d(x^{(l)}, \mu)^2,
\end{equation}
\begin{equation}
    \delta^2_{fr} =  \min_{\mu \in \mathcal{M}} \frac{1}{t}\sum_{l=1}^{t} d(x^{(l)}, \mu)^2.
\end{equation}

This is a generalization of the Euclidean mean since when $\mathcal{M} = \mathit{R}^n$, the optimization problem in Eq.(\ref{eq:mean}) is 

\begin{equation}\label{eq:general}
\begin{split}
 \frac{1}{t}\sum_{l=1}^{t} d(x^{(l)}, \mu)^2
 &= \frac{1}{t}\sum_{l=1}^{t} ||\mu - x^{(l)}||_2^2 = \frac{1}{t}\sum_{l=1}^{t} \sum_{i=1}^{n}(\mu_i - x_i^{(l)})^2,\\
    & = \sum_{i=1}^{n}\left (\mu_i - \frac{1}{t}\sum_{l=1}^{t}x_i^{(l)}\right)^2 \\
    &+ \sum_{i=1}^{n}\left(\frac{1}{t}\sum_{l=1}^{t}(x_i^{(l)})^2 - \left(\frac{1}{t}\sum_{l=1}^{t}x_i^{(l)}\right)^2\right).
\end{split}
\end{equation}

Minimizing the above problem is the definition of the mean and variance in Euclidean space. Thus, this is a natural generalization of the mean to the non-Euclidean space. However, the general formulation of Fr\'echet mean requires an argmin operation and offers no closed form solution, thus both computation and differentiation are problematic. Inspired by work~\cite{gould2016differentiating}, they provided its generalization which allows to differentiate the argmin operation on the manifold. Therefore, they provided their closed form of the Fr\'echet mean.

This can be applied to both Lorentz model and Poincar\'e models and as illustrated in the experimental part~\cite{lou2020differentiating}, this method is much more efficient than the previous Riemannian Gradient Descent (RGD) approach~\cite{udriste2013convex} and against the Karcher Flow algorithm~\cite{karcher1977riemannian}.

\subsection{Concatenation and Split Operations}
Concatenation and split are commonly used in current deep neural networks. They are crucial for feature fusion in multimodal learning, operations like GCN filters and even for computing the correlation in attention mechanism~\cite{vaswani2017attention}. 

As previously described operations, the concatenation and split can also be easily obtained by using the tangent space. Specifically, for an $n$-dimensional feature embedding $x \in \mathbb{B}^n$ in the hyperbolic space, it can be split into $N$ feature representations $\mathit{V}$, 
\begin{equation}
    \mathit{V} = \{v_1 \in \mathbb{R}^{n_1},\ldots,v_N \in \mathbb{R}^{n_N} \} = \text{Log}_0 (x)
\end{equation}
subject to $\sum_{i=1}^{N}n_i = n $. Then, the tangent vector can be mapped to the hyperbolic space using the exponential map. Likewise, for $N$ parts feature representation $\mathit{V}$ in the hyperbolic space, the tangent space can also be used to perform concatenation, which is 
\begin{equation}
    x = \text{Exp}_0 (\text{Log}_0 (v_1)|\text{Log}_0 (v_2),...,|\text{Log}_0 (v_N))
\end{equation}
where the $|$ denotes the concatenation operation in the tangent space and here $v_i$ represents one feature in the hyperbolic space. Therefore, concatenation is the inverse function of split.

However, as pointed out by work~\cite{shimizu2021hyperbolic}, merely splitting the coordinates will lower the norm of the output gyrovectors, which will limit the representational power. Therefore, work~\cite{shimizu2021hyperbolic} proposed a $\beta$-split and $\beta$-concatenation, as an analogy to the generalization criterion in Euclidean neural networks~\cite{he2015delving}.

The $\beta$-split and $\beta$-concatenation provided by~\cite{shimizu2021hyperbolic} introduce a scalar coefficient $\beta_n = B(\frac{n}{2}, \frac{1}{n})$, where $B$ is the Beta distribution. With this scalar coefficient, the tangent vectors are scaled before being projected back to the hyperbolic space. Therefore $\beta$-split is
\begin{equation}
    \mathit{V} = \{\text{Exp}_0 (\beta_{n_1} \beta_{n}^{-1} v_1) ,...,\text{Exp}_0 ( \beta_{n_N}\beta_n^{-1}v_N) \}.
\end{equation}

Therefore, for the $\beta$-concatenation,  
\begin{equation}
    x = \text{Exp}_0 ( \beta_n \beta_{n_1}^{-1} v_1|\beta_n \beta_{n_2}^{-1} v_1 ,...,|\beta_n \beta_{n_N}^{-1}v_N ).
\end{equation}

Work~\cite{ganea2018hyperbolic} presented another way to perform vector concatenation in the hyperbolic space, which introduces linear projection functions based on M\"{o}bius transformations. Specifically, for a set of hyperbolic representations $ \{v_1 \in \mathbb{B}^{n_1},...,v_N \in \mathbb{B}^{n_N} \}$, a group of projection function $ \{M_1 \in \mathbb{B}^{n,n_1},...,M_N \in \mathbb{B}^{n,n_N} \}$ is introduced. Then the concatenated result is :
\begin{equation}
    x = M_1\otimes v_1 \oplus,...,\oplus M_N \otimes v_N.
\end{equation}

However, compared to the previous $\beta$ methods, this one applies the M\"{o}bius transformations (addition and multiplication) many times, as mentioned by~\cite{shimizu2021hyperbolic}, which incurs a heavy computational cost and an unbalanced priority in each input sub-gyrovector.

\subsection{Convolutional Neural Network Operations}

The operations of Convolutional Neural Network (CNN) are fundamental elements for various machine learning tasks, especially for computer vision applications. Many neural network operations have been generalized to hyperbolic spaces, but there is limited research about convolutional layers in this space. Hyperbolic image embedding~\cite{khrulkov2020hyperbolic} proposed to address the common computer vision tasks, e.g., image classification and person re-identification, using hyperbolic geometry. However, the feature representations are still learnt by Euclidean encoder, only the decision hyperplanes are established in the hyperbolic space.  Thus, the authors did not generalize CNN to hyperbolic space. 

Basically, the generalization of CNN can also be simply conducted by using the tangent space. However, whether this approximation works or not is hard to tell. Besides, as pointed out by~\cite{liu2019hyperbolic}, stacking multiple CNNs in the tangent space may collapse to a vanilla Euclidean CNN. In addition, the advantages of hyperbolic geometry may not be well adapted if only using the tangent space. Work~\cite{shimizu2021hyperbolic} provided a novel method to bridge this gap. By using the $\beta$-concatenation and the Poincar\'e Fully connected (FC) layer, the authors presented a method to build the convolutional layer. In particular, given a C-channel input tensor $x \in \mathbb{B}^{C\times W \times H}$ on the Poincar\'e ball, for each of the 
$W \times H$ feature pixels, the representations in the reception field of a convolutional filter with size K are concatenated into a single vector $x \in \mathbb{B}^{nK}$, using the $\beta$-concatenation. Then naturally, a Poincar\'e FC layer, which will be detailed in Sec.~\ref{sec:FC}, can be employed to transfer the feature on the manifold. Let $C^{'}$ be the output channels of the CNN layer, then there will be $C^{'}$ groups of such transformations.

As mentioned before, there is limited research about CNN in the hyperbolic space. The only version provided by work~\cite{shimizu2021hyperbolic} can also be further improved in terms of convolutional strategy and computational efficiency.

\subsection{Recurrent Neural Network Operations}

Recurrent neural networks (RNNs)~\cite{hochreiter1997long,graves2013speech} are one important category of deep neural networks, which allow previous outputs to be used as inputs, thus providing the ability to exhibit temporal behavior. RNNs are commonly utilized in sequence learning tasks. Formally, a RNN can be defined by
\begin{equation}
    h_{t+1} = \delta (\mathcal{W}h_t + \mathcal{U} x_t + b),
\end{equation}
where $h_{t+1}$ is the hidden state of next step, which is updated using current hidden state $h_t$ and input $x_t$. $\delta$ is a non-linear function. The $\mathcal{W}$ and $\mathcal{U}$ are learnable parameters, and $b$ is the corresponding bias. Work~\cite{ganea2018hyperbolic} generalizes the RNN to the hyperbolic space, leveraging the M\"{o}bius operations in Gyrovector space. The RNN in hyperbolic space can be defined by
\begin{equation}
    h_{t+1} = \delta^\otimes (\mathcal{W} \otimes h_t + \mathcal{U} \otimes x_t \oplus b),
\end{equation}
where $\oplus$ and $\otimes$ are the generalization of original $+$ and $\times$ in gyrovector space, as defined in Section~\ref{Sec2}. The authors also extended the same idea into the Gated recurrent unit (GRU) architecture~\cite{cholearning2014GRU}, with the same strategy.

Existing works are limited to the Poincar\'e model with the corresponding operations defined in the Gyrovector space.  However, these kinds of operations are always costly when compared to the Euclidean counterpart. Future works can explore more efficient ways and also extend to other hyperbolic models, like Lorentz model.


\subsection{Activation function}
Activation function is one of the most important components for deep leaning, which provides a non-linear projection of the feed-in features such that more valuable semantic representation can be learnt. As conducted in works~\cite{liu2019hyperbolic, bachmann2020constant}, one can directly apply the \textit{manifold preserving} non-linearity on a manifold, which means it is applied after the exponential map if there is one. However, manifold preserving activation functions are different for different manifolds. As mentioned by~\cite{liu2019hyperbolic}, activation functions like ReLU~\cite{nair2010rectified} and leaky ReLU work as a norm decreasing function in the Poincar\'e model while this is not true for Lorentz model, since the origin in Poincar\'e model is the pole vector in Lorentz model. Thus, to use these activation functions in Lorentz model, one has to map between these two hyperbolic models.

Ganea~\etal~proposed a M\"obius version of projection function~\cite{ganea2018hyperbolic}, which can also be employed to activation functions. In particular, for a function $f: \mathbb{R}^n \rightarrow \mathbb{R}^m$, its M\"obius version $f^\otimes$ is defined as 
\begin{equation}
    f^\otimes (x) = \text{Exp}_0(f(\text{Log}_0(x))).
\end{equation}

The authors utilized the tangent space of the origin to perform the function $f$. This function $f$ can also be the activation function, in which case, the input dimension $n$ is equal to the output dimension $m$. Following the same idea, work~\cite{chami2019hyperbolic} provided a similar activation function for graph convolutional networks. The only difference is that they consider the curvatures of different layers. Thus, the logarithmic map and exponential map are defined at the point origins in the manifold with different curvatures.

There is also work~\cite{shimizu2021hyperbolic} that removes the activation functions, since the authors thought that the operation on the manifold itself is a non-linear operation, which obviates the need for activation functions.

\subsection{Batch Normalization}

Batch Normalization (BN)~\cite{ioffe2015batch} limits the internal covariate shift by normalizing the activations of each layer. Thus it is commonly used to speed up the training procedure of neural networks, as well as to make the training process more stable. The basic idea behind batch normalization is normalizing of the feature representations by re-centering and re-scaling. Specifically, given a batch of $m$ datapoints, The BN algorithm will first compute the mean $\mu$ of this batch. Based on $\mu$, the mini-batch variance $\sigma$ is also computed. Then, two learnable parameters are introduced, which are the scale parameter $\gamma$ and the shift parameter $\beta$. The input activation x is then re-centered and re-scaled, which is $y = \gamma x + \beta$. Therefore, theoretically, this BN operation can be easily generalized to the manifold via transferring to the tangent space. 

Work~\cite{lou2020differentiating} provided an alternative based on Fr\'echet mean~\cite{Frechetmean1948}. In particular, the authors formulated the Riemannian extension of the standard Euclidean Batch Normalization by a differentiable Fr\'echet mean, as described in Sec.~\ref{sec:mean}.

As far as we know, this is the only work for hyperbolic batch normalization. Works for another normalization layers, e.g., Group normalization~\cite{wu2018group} and instance normalization~\cite{ulyanov2016instance}, have not appeared yet. Work~\cite{shimizu2021hyperbolic} pointed out in their hyperbolic neural networks, they removed the activation functions since the operation on the manifold itself has non-linearity. However, whether a hyperbolic generalization of the normalization methods is needed for the bounded manifold (like Poincar\'e model) or not is not clear yet. 

\subsection{Classifiers and Multiclass Logistic Regression}




In the context of deep learning, Multiclass logistic regression (MLR) or softmax regression is commonly used to perform multi-class classification in Euclidean space. Formally, given K classes, MLR is an operation used to predict the probabilities of each class $k \in \{1,2,3,...,K\}$ based on the input representation x, which is
\begin{equation}
    p(y=k,x) \propto \text{exp~}(<a_k, x>-b)
\end{equation}
where $a_k \in \mathbb{R}^n$ denotes the normal vector and $b \in \mathbb{R}$ is the scalar shift. Then, the decision hyperplane determined by $a \in \mathbb{R}\backslash \{\textbf{0}\}$ and $b$ is defined by $\mathit{H}_{a,b} = \{x\in \mathbb{R}^n : <a,x>-b=0 \}$. Note that $ \text{exp~}$ is the exponential function, not the manifold map function $ \text{Exp~}$. According to~\cite{lebanon2004hyperplane}, the MLR can be reformulated as 
\begin{equation}
     p(y=k,x) \propto \text{exp~}( sign(<a_k, x>-b)||a_k||d(x, \mathit{H}_{a_k,b_k}))
\end{equation}
where $d(x, \mathit{H}_{a_k,b_k})$ is the Euclidean distance of x from the hyperplane $\mathit{H}_{a_k,b_k})$. To further generalize it to the hyperbolic space, work~\cite{ganea2018hyperbolic} re-parameterized the scalar term $b \in \mathbb{R} $ with a new set of parameters $p_k \in \mathbb{R}^n$, by which they reformulated the hyperplane: $\hat{\mathit{H}}_{a,p} = \{x\in \mathbb{R}^n : <a,x-p> =0 \}$, and $\hat{\mathit{H}}_{a,p} = \hat{\mathit{H}}_{a,<a,p>}$. In this way, the MLR is rewritten as
\begin{equation}
     p(y=k,x) \propto \text{exp~}( sign(<a_k, x-p_k>)||a_k||d(x, \hat{\mathit{H}}_{a_k,p_k})).
\end{equation}

Then, the definition of the hyperbolic setting is simply achieved by replacing the addition $+$ with M\"{o}bius addition $\oplus$. 

However, this causes an undesirable increase in the parameters from $ n + 1$ to 2n in each class k. As pointed out by~\cite{shimizu2021hyperbolic}, there is no need to introduce countless choices of $p_k$ to determine the same discriminative hyperplane. Instead, they introduced another scalar parameter $r_k \in \mathbb{R}$ such that the bias vector $q_{a_k,r_k} = r_k \frac{a_k}{||a_k||}$ parallels to the normal vector $a_k$. That is 
\begin{equation}
    \hat{\mathit{H}}_{a_k,r_k} = \{x\in \mathbb{R}^n | <a_k, -q_{a_k,r_k}+x> =0\} = {\mathit{H}}_{a_k,r_k||a_k||}).
\end{equation}
Based on this, the MLR is reformulated as
\begin{equation}
     p(y=k,x) \propto exp( sign(<a_k, -q_{a_k,r_k}+x>)||a_k||d(x, \hat{\mathit{H}}_{a_k,r_k})).
\end{equation}
In addition to the MLRs in hyperbolic space, there are also works constructing classifiers in this space. Work~\cite{cho2019large} introduced a hyperbolic formulation of support vector machine classifier (H-SVM). In particular, the Euclidean SVM is trying to optimize a hinge loss function such that the geometric margin can be maximized. Further, for linear SVM classifiers, the maximum-margin problem can be optimized by 
\begin{equation}
\begin{split}
    W^* &= \argmin_w \frac{1}{2}||w||^2, \\
    & \text{subject to}~ y(w^Tx) \geq 1,
\end{split}
\end{equation}
where $(x, y)$ is the training pair, and $w$ is the linear model. Analogous to the Euclidean SVM, they considered a set of decision functions (linear separators) that lead to geodesic decision boundaries in the hyperbolic space (Lorentz model), which is $\mathcal{H}=\{h_w: w\in \mathbb{R}^{n+1}, <w,w>_{\mathbb{L}}<0\}$, where $h_w(x)=1$ if $<w,x>_{\mathbb{L}} >0$, otherwise $h_w(x)=-1$. Here, $<w,w>_{\mathbb{L}}<0$ ensures that the intersection of the decision hyperplane $h_w$ and the manifold $\mathbb{L}^d$ is not empty. The corresponding decision boundaries are the n-dimensional hyperplanes which are given by $<w,x>_{\mathbb{L}} = 0$. They further derived the closed-form of the minimum hyperbolic distance from a data point x to the decision boundaries, which is 
\begin{equation}
    d_{\mathcal{H}}(x) = \text{~arsinh~} \left( \frac{<w,x>_{\mathbb{L}}}{\sqrt{-<w,w>_{\mathbb{L}}}}\right).
\end{equation}
With this distance, they provided the maximum-margin problem in hyperbolic space, which is
\begin{equation}
    \begin{split}
        \argmax \min yh_w (x) \textit{inf}d_{\mathcal{H}}(x),\\
        = \argmax \min \textit{inf} \text{~arsinh~} \left( \frac{ y<w,x>_{\mathbb{L}}}{\sqrt{-<w,w>_{\mathbb{L}}}}\right).
    \end{split}
\end{equation}
Since this distance is scale-invariant, we have $y<w,x>_{\mathbb{L}} \geq 1$. Therefore, the optimization problem now simply maximizes  $\text{~arsinh~} \left( \frac{1}{\sqrt{-<w,w>_{\mathbb{L}}}}\right)$, which is also equivalent to minimizing $\frac{1}{2} <w,w>_{\mathbb{L}}$ subject to the above constraint. Therefore, the linear SVM in the hyperbolic space can be obtained by optimizing 
\begin{equation}
\begin{split}
    W^* = \argmin_w -\frac{1}{2}<w,w>_{\mathbb{L}}, \\
    \textit{subject to}~ y<w,x>_{\mathbb{L}} \geq 1, <w,w>_{\mathbb{L}} < 0.
\end{split}
\end{equation}

The authors also generalized this to a nonlinear case. by extending Euclidean kernel SVM to hyperbolic space.

Sharing the similar idea, work~\cite{weber2020robust} provided theoretical guarantees for learning a SVM in hyperbolic space. Besides, an efficient strategy is introduced to learn a large-margin hyperplane, by the injection of adversarial examples. Specifically, this hyperbolic SVM is also constructed in the Lorentz model. The author mentioned that minimizing this empirical risk via gradient descent requires a large number of iterations, which is not efficient. They alleviated this problem by enriching the training set with adversarial examples. In this way, an optimization function with robust loss is introduced, which is  
\begin{equation}
\begin{split}
  \mathcal{L}_{rob}(w;x,y) = \mathbb{E}~ l_{rob}(x,y;w),\\
  l_{rob}(x,y;w) = \max_{z\in \mathbb{L}: d_{\mathbb{L}}(z,x) \leq \alpha}l(z,y;w),
\end{split}
\end{equation}
where they introduced the adversarial examples to train the SVM model, by perturbing a given input feature x on the hyperbolic manifold. Here, the adversarial example $\hat{x}$ of input sample $x$ is defined as, within a budget $\alpha$, an example that leads to misclassification. Formally, 
\begin{equation}
    \hat{x} = \argmin_{d_{\mathbb{L}}(z,x) \leq \alpha} y <w,z>_{\mathbb{L}}.
\end{equation}

Then the SVM is updated with the gradient-based method on both the original dataset and the corresponding adversarial examples.

\subsection{Fully-Connected Layers}\label{sec:FC}

Fully-Connected layer (FC), or linear transform layer, defined as $Y = \mathit{A}x + b$, is also one important component of deep neural networks, in which all inputs from one layer are connected to every activation unit of the next layer. With analogy to Euclidean FC layers, works~\cite{gulcehre2018hyperbolic,shimizu2021hyperbolic} generalized it to the hyperbolic space. In work~\cite{gulcehre2018hyperbolic}, fully-connected layers~\cite{shimizu2021hyperbolic} are constructed by Matrix-vector multiplication
\begin{equation}
    Y = A\otimes x \oplus b = \text{Exp}_0 (A \text{Log}_0 (x)) \oplus b.
\end{equation}
Here, the bias translations can be further conducted by M\"{o}bius translation, which first map the bias to the tangent space of origin and then parallel transport it to the tangent space of the addend, finally map back the result to the manifold. Formally, it is defined as
\begin{equation}
    x \oplus b = \text{Exp}_0 (\mathcal{PT}_{0\rightarrow x}(\text{Log}_0 (b))) = \text{Exp}_x (\frac{\lambda_0}{\lambda_x}\text{Log}_0 (b))
\end{equation}

However, work~\cite{shimizu2021hyperbolic} pointed out that with the M\"{o}bius translation, such a surface is no longer a hyperbolic hyperplane. Besides, the basic shape of the contour surfaces is determined since the norm of each row vector $a_k$ and bias b contribute to the total scale and shift. To deal with this problem, work~\cite{shimizu2021hyperbolic} provided a new Poincar\'e FC layer based on the linear transformation $\mathit{v}(x)$ mentioned in MLR. They argue that the circular reference between $a_k$ and $q_{a_k,r_k}$ can be unraveled by considering the tangent vector at the origin, $z_k \in \mathcal{T}_0 \mathbb{B}^n$, from which $a_k$ is parallel transported. In this way, they derived a new formulation of the linear transformation, which is 
\begin{equation}
    \mathit{v}_k(x) = 2||z_k||\text{~arsinh~}(\lambda_x <x, \frac{z_k}{||z_k||}> \text{~cosh~}(2r_k)-(\lambda_x - 1)\text{~sinh~}(2r_k)).
\end{equation}

Based on this, they provided their FC layer, which is 
\begin{equation}
    Y = \frac{w}{1+\sqrt{1+||w||^2}},  
\end{equation}
 {where~} $w = (\text{~sinh~}(\mathit{v}_k (x)))$ and the $z_k$ is the generalization of the parameter $a_k$ in A.

\subsection{Optimization}\label{sec:optim}

Stochastic gradient-based (SGD) optimization algorithms are of major importance for the optimization of deep neural networks. Currently, well-developed first order methods include Adagrad~\cite{duchi2011adaptive}, Adadelta~\cite{zeiler2012adadelta}, Adam~\cite{kingma2015adam} or its recent updated one AMSGrad~\cite{reddi2019convergence}. However, all of these algorithms are designed to optimize parameters living in Euclidean space and none of them allows the optimization for non-Euclidean geometries, e.g., hyperbolic space.  

Consider performing an SGD update of the form
\begin{equation}
    W_{t+1} = W_{t} - \alpha g_{t}
\end{equation}
where $g_{t}$ denotes the gradient of a smooth objective and $\alpha > 0$ is the step-size.

Adagrad~\cite{duchi2011adaptive} improves the performance by accumulating the historical gradients,
\begin{equation}
    W_{t+1}^i = W_{t}^i - \frac{\alpha g_{t}^i}{\sqrt{\sum_{k=1}^{t}(g_{k}^i)^2}}.
\end{equation}
Note that this is \textbf{coordinate-wise} updating. Accumulation of historical gradients can be particularly helpful when the gradients are sparse, but updating with historical gradients is computational inefficient. To deal with this issue, the well-known Adam optimizer further modified it by adding a momentum term $m_t =\beta_1 m_{t-1}+ (1-\beta_1) g_{t}$ and an adaptive term $v_t^i = \beta_2v_{t-1}^i+ (1-\beta_2) (g_t^i)^2$. Therefore, the Adam update is given by
\begin{equation}
    W_{t+1}^i = W_{t}^i - \frac{\alpha m_{t}^i}{\sqrt{v_{t}^i}}.
\end{equation}

In terms of the optimizer on Riemannian manifolds, one pioneer work for stochastic optimization on the Riemannian manifold should be the Riemannian stochastic gradient descent (RSGD), provided by Bonnabel \etal ~\cite{bonnabel2013stochastic}. They pointed out that for the standard stochastic gradient in $\mathbb{R}^n$, seeking the matrix with certain rank, which best approximates the updated matrix, can be numerically costly, especially for very large parameter matrix. To enforce the rank constraint, a more natural way is to endow the parameter space with a Riemannian metric and perform a gradient step within the manifold of fixed-rank matrices. To address this issue, they proposed to replace the usual update in SGD using an exponential map ($\text{Exp}$) with the following update
\begin{equation}
     W_{t+1} = \text{Exp}_{W_{t}}(- \alpha g_{t})
\end{equation}
where $g_{t} \in \mathcal{T}_{W_{t}}\mathcal{M}$ denotes the Riemannian gradient of the objective at point $W_{t}$. The provided algorithm is completely intrinsic, which does not depend on a specific embedding of the manifold or on the choice of local coordinates. For manifolds where $\text{Exp}$ is not known in closed-form, it is also common to replace it with a first order approximation. The following work from~\cite{zhang2016first} provided a global complexity analysis for first-order algorithms for general g-convex optimization, under geodesic convexity.

On the top of RSGD, Work~\cite{nickel2017poincare} derived the gradient for Poincar\'e model. They utilized the retraction operation $\mathcal{R}_w(v) = w+v$ as the approximation of the exponential map,$\text{Exp}$. Furthermore, a projection, which just normalizes the embeddings with a norm bigger than one, is utilized to ensure the embeddings remain within the Poincar\'e model. Specifically,
\begin{equation}
     W_{t+1} = proj \left(W_{t} - \eta \frac{(1-||W_{t}||^2)^2}{4} \triangledown_{E} \right),
\end{equation}
where $\triangledown_{E}$ is the Euclidean gradient, and $\eta $ is the learning rate. The normalization function is $ proj(x) = x/||x||$ if $||x|| \geq 1$, otherwise $proj(x) = x$. 
Work~\cite{chamberlain2019scalable} shares the same idea and extends it to the Lorentz model. In particular, there are three steps: firstly, the Euclidean gradients of the objective are re-scaled by the inverse Minkowski metric. Then, the gradient is projected from the ambient space to the tangent space. Finally, the points on the manifold are updated through the exponential map with learning rate.

One of the current challenges of generalizing the adaptivity of these optimization methods to non-Euclidean space is that Riemannian manifold does not provide an intrinsic coordinate system, which leads to the meaningless of sparsity and coordinate update of the gradient. Riemannian manifold only allows to work in certain local coordinate systems, which is called charts. Since several different charts can be defined around each point $x \in \mathcal{M}$, it is hard to say that all the optimizers above can be extended to $\mathbb{R}$ in an intrinsic manner (coordinate-free). Work~\cite{bcigneul2019riemannian} addressed this problem using the cartesian product manifold, as well, provided a generalization of the well-known optimizers, e.g., Adagrad and Adam, with corresponding convergence proofs. 

As pointed out by~\cite{bcigneul2019riemannian}, one solution can be fixing a canonical coordinate system in the tangent space and then parallel transporting it along the optimization trajectory. However, general Riemannian manifold depends on both the chosen path and the curvature, which will give to the parallel transport a rotational component (holonomy). This will break the gradient sparsity and thus hence the benefit of adaptivity. As provided by work~\cite{bcigneul2019riemannian}, an $n$-dimensional manifold $\mathcal{M}$ can be represented by a Cartesian product of $n$ manifolds, which means $\mathcal{M} = \mathcal{M}_1 \times...\times \mathcal{M}_n\times$.  Based on this, the authors provided the Riemannian Adagrad, which is defined by
\begin{equation}
     W_{t+1} = \text{Exp}_{W_{t}}(- \alpha g_{t})
\end{equation}
\begin{equation}
    W_{t+1}^i = \text{Exp}_{W_{t}^i} \left( - \frac{\alpha g_{t}^i}{ \sqrt{\sum_{k=1}^{t}||g_{k}^i||_{x_{k}^i}^2}} \right),
\end{equation}
where the $||g_{k}^i||_{x_{k}^i}^2= \mathfrak{g}(g_{k}^i, g_{k}^i)$ is Riemannian norm. They further extended to Riemannian Adam by introducing the momentum term and an adaptive term.  

Another work worth paying attention to is the Geoopt~\cite{geoopt2020kochurov}, which provided a Riemannian Optimization in PyTorch~\cite{paszke2019pytorch}. Geoopt provides a standard Manifold interface for Riemannian Optimization. It does not only support basic Riemannian SGD, but advanced adaptive optimization algorithms as well.

\section{Neural Networks in Hyperbolic space}\label{Sec5}

\begin{table*}[thbp!]\footnotesize
\begin{center}
\caption{ \small{Summary of the commonly used Poincar\'e model and Lorentz model in hyperbolic spaces.Here, we constrain the model to the unit space (Curvature is fixed to one)}, one can also extend it to other negative curvature}
\label{tab:summeryBandL}
\begin{center}
\begin{tabular}{ l c c } 
\toprule
&\textbf{Poincar\'e Ball} & \textbf{Lorentz Model}  \\
\hline
\textbf{Manifold} & ($\mathbb{B}$, $\mathfrak{g}_x^\mathbb{B}$) & ($\mathbb{L}$, $\mathfrak{g}_x^\mathbb{L}$)  \\ 
\hline
\textbf{Metric} & $g_x^\mathbb{B} = (\frac{2}{1-||x||^{2}})^{2}g^{\mathit{E}}$ & $g_x^\mathbb{L} = I ~\text{except that}~ I_{00} = -1$ \\ 
\hline
\textbf{Distance} & $d(x,y) = \text{~arcosh~}(1+2\frac{||x-y||^2}{(1-||x||^2)(1-||y||^2)})$ & $d(x,y) = \text{~arcosh~}(-<x,y>_{\mathbb{L}})$  \\ 
\hline
\textbf{Parallel Transport} & $\mathit{PT}_{0\rightarrow x} (v) = \text{Log}_x (x \oplus \text{Exp}_0 (v)) = \frac{\lambda_0}{\lambda_x}v$  & $\mathit{PT}_{ 0\rightarrow x} (v) = v- \frac{<\text{Log}_0 (x), v>_{\mathbb{L}}}{d_{\mathbb{L}}(0,x)^2} (\text{Log}_0 (x)- \text{Log}_x (0))$  \\ 
\hline
\textbf{Exponential map} & $\text{Exp}_x(v) = x \oplus (\text{~tanh~}(\frac{\lambda_x||v||}{2})\frac{v}{||v||})$ & $\text{Exp}_x(v) = \text{~cosh~}(||v||_{\mathbb{L}})+\text{~sinh~}(||v||_{\mathbb{L}})\frac{v}{||v||_{\mathbb{L}}}$  \\ 
\hline
\textbf{Logarithmic map} & $\text{Log}_x(y) = \frac{2}{\lambda_x}\text{~artanh~}(||-x\oplus y||)\frac{-x\oplus y}{||-x\oplus y||}$ & $\text{Log}_x(y) = d_{\mathbb{L}}(x,y)\frac{y+<x,y>_{\mathbb{L} }x}{||y+<x,y>_{\mathbb{L}}x||_{\mathbb{L}}}$ \\ 
\bottomrule
\end{tabular}
\end{center}
\end{center}
\end{table*}


 
 

In this section, we will describe the generalization of neural networks into hyperbolic space. We try our best to collect and summarize all the advanced hyperbolic machine learning methods, as illustrated in Table~\ref{tab:summaryNN}.

 One can find that most of the works are from the prominent machine learning conferences like, NeurIPS, ICML, and ICLR. And more and more institutions are getting into this potential research field. From the method perspective, the Poincar\'e model and Lorentz model are most-welcomed hyperbolic models for generalizing neural networks. At the same time, the former one (Poincar\'e model) is dominated in the hyperbolic deep neural networks.  The corresponding important components of these two models are summarized in Table~\ref{tab:summeryBandL}. In the following part, we detail different kinds of hyperbolic architectures, including hyperbolic embedding, hyperbolic clustering, hyperbolic attention networks, hyperbolic graph neural networks, hyperbolic normalizing flow, hyperbolic variational auto-encoder, and hyperbolic neural networks with mixed geometries.

\subsection{Hyperbolic Embeddings}

As far as we know, work~\cite{nickel2017poincare} is the first to propose to learn an embedding using Poincar\'e model, while considering the latent hierarchical structures. They also proved that Poincar\'e embeddings can outperform Euclidean embeddings significantly on data with latent hierarchies, both in terms of representation capacity and in terms of generalization ability. Specifically, the Poincar\'e embedding is trying to find embeddings $\Theta = \{\theta_i\}_{i=1}^{n}, \theta \in \mathbb{B}^d, ||\theta_i|| < 1$ in the unit d-dimensional Poincar\'e ball for a set of symbols with size of n. Therefore, the optimization problem can be framed as $\Theta* = \argmin_{\Theta} \mathcal{L}(\Theta)$, where $\mathcal{L}(\cdot)$ is a task-related loss function. For instance, in the hierarchy embedding task, the loss function over the entire dataset $\mathcal{D}$ can be represented as
\begin{equation}
    \mathcal{L}(\Theta) = \sum_{ (u,v)\in \mathcal{D}} log \frac{e^{-d_{\mathbb{B}}(u,v)}}{\sum_{v^{'} \in \mathcal{N}(u)} e^{-d_{\mathbb{B}}(u,v^{'})}},
\end{equation}
where $\mathcal{N}(u)$ denotes a set of negative examples for u. This loss function encourages related objects to be closer to each other than objects without an obvious relationship. They further optimized the embedding utilizing the RSGD with the exponential map, which is a scaled version of the Euclidean gradient, as describe in Sec.~\ref{sec:optim}.

Based on~\cite{nickel2017poincare}, work~\cite{dhingra2018embedding} also proposed text and sentences embedding with Poincar\'e model. However, to benefit from the advanced optimization techniques like Adam, the authors proposed to re-parametrize the Poincar\'e embeddings such that the projection step is not required. On top of existing encoder architectures, e.g., LSTMs or feed-forward networks, a reparameterization technique is introduced to map the output of the encoder to the Poincar\'e ball. Specifically, the re-parameterized embedding is defined as 
\begin{equation}
    \Theta = \delta(\phi_{norm}(h)) \frac{\phi_{dir}(h)}{||\phi_{dir}(h)||},
\end{equation}
where $h$ represents the hidden representation of encoder, and $\delta$ denotes the sigmoid function. The function $\phi_{dir}$ is used to compute a direction vector $\phi_{dir}: \mathbb{R}^d \rightarrow \mathbb{R}^d$. Function $\phi_{norm}$ is a norm function $\phi_{norm}: \mathbb{R}^d \rightarrow \mathbb{R}$. In this way, the authors mapped the encoder embeddings to the Poincar\'e ball and the Adam is introduced to optimize the parameters of the encoder.

Work~\cite{ganea2018hyperbolic} also contributes to improve the Poincar\'e embeddings from work~\cite{nickel2017poincare}, in the task of entailment analysis. The authors pointed out that most of the embedding points with the method~\cite{nickel2017poincare} collapse on the boundary of Poincar\'e ball. Besides, there is also no advantage for the method to deal with more complicated connections like asymmetric relations in directed acyclic graphs.  To address these issues, the authors presented hyperbolic entailment cones for learning hierarchical embeddings. In fact, with the same motivation, the order embeddings method~\cite{vendrov2016order} explicitly models the partial order induced by entailment relations between embedded objects, in the Euclidean space. However, the linearly growing of this method will unavoidably lead to heavy intersections, causing most points to end up undesirably close to each other. Generalizing and improving over the idea of order embeddings, this work~\cite{ganea2018hyperbolic} viewed hierarchical relations as partial orders defined using a family of nested geodesically convex cones, in the hyperbolic space. 

Specifically, in a vector space, a convex cone S is a set that is closed under non-negative linear combinations, which is for vectors $v_1, v_2 \in S$, then $\alpha v_1 + \beta v_2 \in S, \forall \alpha, \beta > 0$. Since the cones are defined in the vector space, the authors proposed to build the hyperbolic cones using the exponential map, which leads to the definition of \textit{S}-cone at point x
\begin{equation}
    \mathfrak{S}_x = \text{Exp}_x(S), S \in \mathcal{T}_x \mathcal{M}.
\end{equation}
To avoid heavy cone intersections and scale exponentially with the space dimension, the authors further constructed the angular cones in the Poincar\'e ball. To achieve this, they introduced the so-called cone aperture functions $\phi(x)$ such that the angular cones $\mathfrak{S}_x^{\phi(x)}$ follow four intuitive properties, including axial symmetry, rotation invariant, continuous of cone aperture function, and the transitivity of nested angular cones. With all of this the authors provided a closed form of the angular cones, which are
\begin{equation}
    \mathfrak{S}_x^{\phi(x)} = \left\{ (\pi- \angle Oxy) \leq \text{arsin~} (K \frac{1-||x||^2}{||x||}) \right\},
\end{equation}
where $O$ in angle $\angle Oxy$ is origin. K is a constant. Since they also provided a closed form expression for the exponential function $\textit{exp}$, they optimize the objective using fully Riemannian optimization, instead of using the first-order approximation as work~\cite{nickel2017poincare} did. 

Work~\cite{de2018representation} presented a combinatorial construction approach to embed trees into hyperbolic spaces without performing optimization. The resulting mechanism is analyzed to obtain dimensionality-precision trade-offs.  Based on~\cite{sarkar2011low} and ~\cite{abraham2007reconstructing}, they first produced an embedding of a graph into a weighted tree, and then embedded that tree into the hyperbolic disk. In particular, Sarkar’s construction can be summarized like this. Consider two nodes a and b in a tree where node b is the parent of a. With the embedding function $f$, they are embedded as $f(a)$ and $f(b)$. And the children of a, $c_1, c_2,..,c_n$, are to be placed. Then, reflect $f(a)$ and $f(b)$ across a geodesic such that they are mapped to the origin and a new point z, respectively. Then, the children are placed at points around a circle and at the same time maximally separated from z. Formally, let $\theta$ be the angle of $z$ from $x$-axis in the plane. For the $i$-th child $c_i$, the projected place $y_i$ is 
\begin{equation}
    y_i = \left( \frac{e^r-1}{e^r+1} \text{cos~}(\theta + \frac{2\pi i}{n}), \frac{e^r-1}{e^r+1}\text{sin~}(\theta + \frac{2\pi i}{n}) \right),
\end{equation}
where $\frac{e^r-1}{e^r+1}$ works as the radius and $r$ is a scaling factor. After getting all places of the children nodes,  the points are reflected back across the geodesic so that all children are at a distance r from the parent $f(a)$. To embed the entire tree, one can place the root at the origin and its children in a circle around it, and recursively place their children until all nodes have been placed. For high-dimensional spaces, the points are embedded into hyperspheres~\cite{conway2013sphere}, instead of circles. 

Next, they extended to the more general case in the hyperbolic space by the Euclidean multidimensional scaling~\cite{cox2008multidimensional}. To answer the question that given the pairwise distances arising from a set of points in hyperbolic space, how to recover the original points, a hyperbolic generalization of the multidimensional scaling (h-MDS) is proposed, deriving a perturbation analysis of this algorithm following Sibson \etal ~\cite{sibson1979studies}.

Work~\cite{tifrea2019poincarGlove} adapted the Glove~\cite{pennington2014glove} algorithm, which is an Euclidean unsupervised method for learning word representations from statistics of word co-occurrences and wanting to geometrically capture the meaning and relations of a word, to learn unsupervised word embeddings in this type of Riemannian manifolds. To this end, they proposed to embed words in a Cartesian product of hyperbolic spaces which they theoretically connected to the Gaussian word embeddings and their Fisher geometry. Note that, in work~\cite{tifrea2019poincarGlove}, Poincar\'e Glove absorbs squared norms into biases and replaces the Euclidean with the Poincar\'e distance to obtain the hyperbolic version of Glove.

While most of the works concentrated on providing new embedding methods in hyperbolic space, very limited ones noticed the numerical instability of the networks. As pointed out by work~\cite{yu2019numerically}, the difficulty is caused by floating-point computation and amplified by the ill-conditioned Riemannian metrics. As points move far away from the origin, the error caused by using floating-point numbers to represent them will be unbounded. Specifically, the Poincar\'e model and Lorentz model are the most commonly used ones when building neural networks in hyperbolic space. However, for the Poincar\'e model, the distance changes rapidly when the points are close to the ball boundary such that it is not well conditioned. While for Lorentz model, it is not bounded such that it will experience large numerical error when the points are far away from the origin.  Therefore, for the representation in the hyperbolic space, it is desirable to find a method that can represent any point with small fixed bounded error in an efficient way. To address this issue, work~\cite{yu2019numerically} presented a method that using the integer-lattice square tiling (or tessellation)~\cite{conway2016symmetries} in the hyperbolic space to construct a constant-error representation. In this way, a tiling-based model for hyperbolic space is provided. We prove that the representation error, the error of computing distances, and the error of computing gradients is bounded by a fixed value that is independent of how far the points are from the origin.

Specifically, the model from~\cite{yu2019numerically} is constructed on top of Lorentz model. Then the Fuchsian groups~\cite{katok1992fuchsian} G and the fundamental domain~\cite{yuncken2011regular,katok1992fuchsian} F were introduced to build the tiling model.  In the 2-dimensional Lorentz model, the isometries can be represented as matrices $A \in \mathbb{R}^{3\times 3}$ operating on $\mathbb{R}^3$ that preserve the Lorentzian scalar product, which is $A^T \mathfrak{g}^L A = \mathfrak{g}^L$. As mentioned by~\cite{kolbe2019tiling,voight2014arithmetic}, any tiling is associated with a discrete group G of orientation preserving isometries of Lorentz plane $\mathbb{L}^2$ that preserve the tiling. Then, the Fuchsian groups~\cite{katok1992fuchsian} are employed which are the discrete subgroups of isometries $\mathbb{L}^2$. Based on G and F, the tiling model is defined as the Riemannian manifold $(\mathcal{T}_l^n, \mathfrak{g}^L)$, where 
\begin{equation}
\begin{split}
        \mathcal{T}_l^n = \{(g,x) \in G\times F: <x,x> = -1\},\\
    d_{lt}((g_x,x),(g_y,y))= \text{~arcosh~}(-x^Tg_x^T\mathfrak{g}^Lg_yy),
\end{split}
\end{equation}
where $(g,x)$ is the ordered pair with $g\in G$,$x \in F$, and $(g,x)$ represents the point $gx$. They further constructed a special kind of Fuchsian groups G with integers such that it can be computed exactly and efficiently. The group is defined as $G = L<g_a,g_b|g_a^6=g_b^6=(g_ag_b)^3=1>L^{-1}$, where
\begin{equation}
    g_a = \begin{bmatrix}
 2& 1 &0 \\ 
 0& 0 &-1 \\ 
 3& 2 & 0
\end{bmatrix} , g_b = \begin{bmatrix}
 2& -1 &0 \\ 
 0& 0 &-1 \\ 
 -3& 2 & 0
\end{bmatrix} ,\text{and~} L = \begin{bmatrix}
 \sqrt{3}& 0 &0 \\ 
 0& 1 &0 \\ 
 0& 0 & 1
\end{bmatrix} 
\end{equation}
It is easy to verify that the group in G is an isometry since $(Lg_a L^{-1})^T\mathfrak{g}^L(Lg_aL^{-1})=\mathfrak{g}^L$, likewise $(Lg_b L^{-1})^T\mathfrak{g}^L(Lg_b L^{-1})=\mathfrak{g}^L$. By this way, they can perform integer arithmetic which is much more efficient. The authors also extended this 2-dimensional model to higher dimensional spaces by either taking the Cartesian product of multiple copies of the tiling model, or constructing honeycombs and tiling from a set of isometries that is not a group.

\subsection{Hyperbolic Cluster Learning}

Hierarchical Clustering (HC)~\cite{cohen2019hierarchical} is a recursive partitioning of a dataset into clusters at an increasingly finer granularity, which is a fundamental problem in data analysis, visualization, and mining the underlying relationships. In hierarchical clustering, constructing a hierarchy over clusters is in the form of a multi-layered tree whose leaves correspond to samples (datapoints of a dataset) and internal nodes correspond to clusters. 

Bottom-up linkage methods, like Single Linkage, Complete Linkage, and Average Linkage~\cite{yim2015hierarchical}, are one of the first suite of algorithms to address the HC problem, which recursively merge similar data-points to form some small clusters and then gradually larger and larger clusters emerge. However, these methods are optimized using heuristic strategies over discrete trees, of which the discreteness restricts them from the benefits of advanced and scalable stochastic optimization. Recently, cost function based methods have increasingly gained attention. The recent article of Dasgupta~\cite{dasgupta2016cost} served as the starting point of this topic. Dasgupta framed similarity-based hierarchical clustering as a combinatorial optimization problem, where a “good” hierarchical clustering is one that minimizes a particular cost function, which is
\begin{equation}\label{eq:costHC}
    \textit{L}(\mathcal{T};w) = \sum_{i,j}w_{ij}|lvs(\mathcal{T}(i \vee j))|,
\end{equation}
where $\mathcal{T}$ is a tree-structured hierarchical clustering. $w_{ij}$ denotes the similarity, $i \vee j$ represents Lowest Common Ancestor (LCA) of node i and j, $\mathcal{T}(i \vee j)$ is a subtree rooted at the LCA $i \vee j$. The $lvs(LCA)$ gets the leaves of the LCA node.  Therefore, Eq.~(\ref{eq:costHC}) encourages similar nodes to be nearby in the same tree structure, which is based on the motivation that a good tree should cluster the data such that similar data-points have LCAs much further from the root than that of the dissimilar data-points.

Nevertheless, as pointed out by~\cite{monath2019gradient}, due to its discrete nature, this objective is not amenable for stochastic gradient methods. To deal with this issue, recently they proposed a related cost function over triples of data points, based on a ranking of candidate configurations of the tree. The basic idea is if all of the three points $i,j,k$ have the same LCA, the cost is the sum of all pairwise similarities between the three points. Otherwise, if two points have a LCA that is deeper in the tree than the LCA of all three, then the cost is the sum of all pairwise similarities between the three points except that of the pair with the deeper LCA. Formally, it is 
\begin{equation}\label{eq:cHC}
    \textit{L}(\mathcal{T};w) = \sum_{i,j,k}[w_{ij}+w_{ij}+w_{ij}-w_{ord}(\mathcal{T};w)]+ 2\sum_{i,j}w_{ij},
\end{equation}
where the $w_{ord}(\mathcal{T};w) = w_{ij}\mathbbm{1}[\{i,j|k\}]+ w_{ik}\mathbbm{1}[\{i,k|j\}]+w_{jk}\mathbbm{1}[\{j,k|i\}]$, is defined through the notion of LCA. The $\mathbbm{1}[]$ is the indicator function which is one if condition satisfied and zero otherwise. It will be zero if the three points have the same LCA. Here, The relation $\{i,j|k\}$ holds if $i\vee j$ is a descent of $i\vee j \vee k$.

However, this cost function is computationally expensive, especially for large-scaled applications. To deal with this issue, gradient-based hyperbolic hierarchical clustering, gHHC~\cite{monath2019gradient}, a geometric heuristic to provide an approximate distribution over LCA, is proposed over continuous representations of tree in the hyperbolic space (Poincar\'e model), based on the observation that Child-parent relationships can be modelled by the distances and norms of the embedded node representations. The authors used the norm of vectors to model depth in the tree, requiring child nodes to have a larger norm than their parents. The root is near the origin of the space and leaves near the edge of the ball. Formally, let $Z=\{z_1, z_2,...,z_k\}, z_i \in \mathbb{B}^d$ represent the node representation in the $d$-dimensional Poincar\'e ball. Then a child-parent dissimilarity function is used to encourage the children have a smaller norm than the parent, which is
\begin{equation}
    d_{cp}(\mathcal{T}_c, \mathcal{T}_p) = d_{\mathbb{B}}(z_c, z_p)(1+max(||z_p||-||z_c||, 0)),
\end{equation}
where $z_c, z_p$ are the hyperbolic embedding of nodes $\mathcal{T}_c, \mathcal{T}_p$. If the norm of the parent node is smaller than the child, then the dissimilarity will just be the distance in hyperbolic space. Otherwise, this dissimilarity will be bigger than the distance. Then this dissimilarity function is used to model a distribution over the tree structure to encode the uncertainty, which is $P_{par}(\mathcal{T}_p|\mathcal{T}_c,Z) \propto exp(-d_{cp}(\mathcal{T}_c, \mathcal{T}_p))$, thus the tree distribution over embedding will be
\begin{equation}
    P_{par}(\mathcal{T}|Z) \propto \prod_{\mathcal{T}_p} \prod_{\mathcal{T}_c \in children(\mathcal{T}_p)} P_{par}(\mathcal{T}_p|\mathcal{T}_c,Z).
\end{equation}




\subsection{Hyperbolic Attention Network}

Attention mechanism~\cite{vaswani2017attention} is one of the currently most attractive research topics. The well-known architectures like the neural Transformer~\cite{vaswani2017attention}, BERT~\cite{devlin2019bert}, hyperbolic attention network~\cite{gulcehre2018hyperbolic}and even the graph attention networks~\cite{velivckovic2017graph, zhang2020hyperbolic,peng2020learning,peng2020mix}. While attention mechanisms have become the de-facto standard for NLP tasks, their momentum has continuously been extended to computer vision applications~\cite{dosovitskiy2020image}. At its core lies the strategy of focusing on the most relevant parts of the input to make decisions. When an attention mechanism is used to compute a representation of a single sequence, it is commonly referred to as self-attention or intra-attention. For each of the input vectors (embeddings), the attention will first create a Query vector, a Key vector, and a Value vector. These three vectors are simply constructed by multiplying the embedding by three learnable projection matrices. The core computation of the attention is the attentive read operation, which calculates a normalized weighted value for each query with corresponding multiple locations (keys). The attentive read is formed as
\begin{equation}
    r(q_i,\{k_j\}_j) = \frac{1}{Z}\sum_{j}\alpha(q_i, k_j)v_{ij},
\end{equation}
where Z is a normalization factor and  $q_i, k_j$ and $v_{ij}$ are the query, key and value, respectively. The function $\alpha(\cdot)$ is used to compute the match score. According to work~\cite{gulcehre2018hyperbolic}, this can be broken down into two parts, e.g., first score matching and then information aggregation. In the transformer model~\cite{vaswani2017attention}, the matching function is $\alpha(q_i, k_j)= exp \left( \frac{1}{\sqrt{d}} <q_i, k_j>\right)$, the value $v_{ij} = v_j$, $Z=\sum_j \alpha(q_i, k_j)$. In the graph attention network~\cite{velivckovic2017graph}, the score matching function computes the score between different graph nodes activations $h_i$ and $h_j$, which is 
\begin{equation}\label{eq:match}
  \alpha_{ij} = f(Wh_i, Wh_j)  
\end{equation}

The authors provided a normalized version using a single layer feed-forward network, which is
\begin{equation}
    \alpha_{ij} = \frac{\text{exp~}(\text{LeakyReLU}(a^T [Wh_i||Wh_j]))}{\sum_{k \in \mathcal{N}_i} \text{exp~}(\text{LeakyReLU}(a^T [Wh_i||Wh_k]))},
\end{equation}
where $||$ denotes the concatenation operation, $a^T$ and W are linear transformations. $k \in \mathcal{N}_i$ represents the neighbors of node i. Then the information is aggregated around the neighbors.

Based on the above mentioned attention mechanism, work~\cite{gulcehre2018hyperbolic} extended it into the hyperbolic space (Lorentz model). Specifically, the Lorentzian distance and the Einstein midpoint method are introduced to conduct the score matching and aggregation while building the hyperbolic space. First, the data representation is organized by a pseudo-polar coordinate, in which an activation $x\in \mathbb{R}^{n+1} = (\textbf{d},r)$ is constructed by a n dimensional normalized angle \textbf{d},$||\textbf{d}||= 1$ and a scalar radius r.  Then, a well-developed map function is introduced to project the activation to the hyperbolic space, which is $\pi((\textbf{d}, r))= (\text{~sinh~}(r)\textbf{d}, \text{~cosh~}(r))$. It is easy to see that the projected point lies in the Lorentz model. Then for hyperbolic matching, the authors took $\alpha (q_i, k_j)= f(-\beta d_{\mathcal{L}(q_i, k_j)}-c)$, in which the negative Lorentzian distance (scaled by $-\beta$ and shifted by $c$) is utilized to measure the correlation. Since there is no natural definition of mean on the manifold, they turn to Einstein midpoint to conduct hyperbolic aggregation. Specifically, the agregated message $m_i$ can be represented as
\begin{equation}
    m_i(\{\alpha_{ij}\}_j, \{v_{ij}\}_j) = \sum_j \left[ \frac{\alpha_{ij} \gamma(v_ij)}{\sum_l \alpha_{il} \gamma(v_il)}\right] v_{ij},
\end{equation}
where $\gamma(v_ij)$ is the Lorentz factor. However, one thing to be noted is that the Einstein midpoint is defined on the Klein model. Since these hyperbolic models are isomorphic, it is very easy to map between them, as described in Sec.~\ref{sec:isom}. On the top of the proposed hyperbolic attention network, the authors further formulated the Hyperbolic Transformer model, which is proved to have the superiority when compared to the Euclidean Transformer.

Based on the Euclidean graph attention network (GAT)~\cite{velivckovic2017graph}, work~\cite{zhang2020hyperbolic} generalizes it to a hyperbolic GAT. The idea is very simple, just replacing the Euclidean operation with M\"obius operations. In particular, the Eq.~(\ref{eq:match}) changes to
\begin{equation}
    \alpha_{ij} = f(W\otimes h_i, W\otimes h_j). 
\end{equation}
They further defined the $f$ function based on the hyperbolic distance. Just as work~\cite{gulcehre2018hyperbolic}, the negative of the distance between nodes is utilized as the matching score. The scores are finally normalized using softmax function, otherwise all the score are negative. The hyperbolic aggregation is simply conducted on the tangent space, as it is done in Euclidean space.

Currently, many works have made an attempt to generalize centroid~\cite{law2019lorentzian,shimizu2021hyperbolic} operations to hyperbolic space, since it has been utilized in many recent attention-based architectures. Work~\cite{shimizu2021hyperbolic} proved that three different kinds of hyperbolic centroids, including the M\"{o}bius gyromidpoint~\cite{ungar2008gyrovector}, Einstein midpoint~\cite{ungar2008gyrovector} and the centroid of the squared Lorentzian distance~\cite{law2019lorentzian},  are the same midpoint operations projected on each manifold and exactly matches each other. Based on this observation, they explored on M\"{o}bius gyromidpoint and generalized it by extending to the entire real value weights (previously, it is defined under the condition of non-negative weights) by regarding a negative weight as an additive inverse operation. So the centroid with real weights $\{\textit{v}_i \in \mathbb{R}\}_{i=1}^{N}$ is 
\begin{equation}
    \mu = [x_i,\textit{v}_i] = \frac{1}{2} \oplus \left( \frac{\sum_{i=1}^{N} \textit{v}_i \lambda x_i}{\sum_{i=1}^{N} ||\textit{v}_i|| \lambda x_i}\right)
\end{equation}
With the above weights, we can introduce how to compute the attention in hyperbolic space.  Given the source and target as sequences of gyrovectors, firstly, the proposed Poincar\'e FC layers, as detailed in Sec.~\ref{sec:FC}, are utilized to construct the queries, keys, and values. Then they are broken down into several parts, utilizing the Poincar\'e $\beta$-splits, in order to build the multi-head attention. Sharing the same idea, the negative distances are also employed to measure the matching scores. Finally, the message from multi-head is aggregated using the proposed Poincar\'e weighted centroid. The authors also built a hyperbolic set Transformer model and compared to its Euclidean counterpart~\cite{lee2019set}. The result shows that the hyperbolic one can at least get equivalent performance, at the same time showing a remarkable stability and consistently converges.

\subsection{Hyperbolic Graph Neural Network}
Graph neural networks (GNNs) is a very hot research topic in the machine learning community, especially in the fields dealing with data processing non-Euclidean structures. Rencently, this is a growing passion of modeling graph in the hyperbolic space~\cite{liu2019hyperbolic,peng2020mix,bachmann2020constant}. A core reason for that is learning hierarchical representations of graphs is easier in the hyperbolic space due to the curvature and the geometrical properties of the hyperbolic space. Such spaces were shown by Gromov to be well suited to represent tree-like structures~\cite{gromov1987hyperbolic} as objects requiring an exponential number of dimensions in Euclidean space can be represented in a polynomial number of dimensions in hyperbolic space.

The success of conventional deep neural networks for regular grid data has inspired generalizations for analysing graphs with graph neural networks (GNN). Originally, GNN was proposed by~\cite{scarselli2008graph,gori2005new} as an approach for learning graph node representations via neural networks. This idea was extended to convolutional neural networks using spectral methods~\cite{bruna2013spectral,henaff2015deep} and spatial methods with iterative aggregation of neighbor representations~\cite{henaff2015deep,kipf2016semi}.

With the development of hyperbolic neural networks and hyperbolic attention networks provided by Ganea \etal~\cite{ganea2018hyperbolic} and Gulcehre \etal~\cite{gulcehre2018hyperbolic} to extend deep learning methods to hyperbolic space, interest has been rising lately towards building graph neural networks in hyperbolic spaces. Before we further introduce the graph neural networks in the hyperbolic space, let us briefly introduce the background of GNN. 

Graph neural networks can be interpreted as performing message passing between nodes. The information propagation on the graph can be defined as
\begin{equation}\label{eq:gnn}
    h_i^{k+1} = \sigma \left( \sum_{j \in \mathit{N}(i)} A_{ij} \mathcal{W}^k h_i^{k}\right),
\end{equation}
where $h_i^{k+1}$ represents the hidden representation of the $i$-th node  at the $(k+1)$-layer, $\mathcal{W}^k$ denotes the weight of the network at k layer. The $A_{ij}$ is the entry of the normalized adjacency matrix $A$. Eq.~(\ref{eq:gnn}) performs the information aggregation around the neighbor nodes $\mathit{N}(i)$ of node i to update the representation of this node. 

Works~\cite{liu2019hyperbolic,peng2020mix} provide a straightforward way to extend the graph neural network to hyperbolic space, using tangent space, since the tangent space of a point on Riemannian manifolds is always a subset of Euclidean space~\cite{liu2019hyperbolic}. 
Work~\cite{liu2019hyperbolic} utilized the logarithmic map $\text{Log}_{x'}$ at a chosen point $x'$, such that the functions with trainable parameters are executed there. Therefore, we get the graph neural operations in a hyperbolic space, which is 
\begin{equation}
    h_i^{k+1} = \sigma ( \text{Exp}_{x'}\left(\sum_{j \in \mathit{N}(i)} A_{ij} \mathcal{W}^k \text{Log}_{x'}(h_i^{k})\right)),
\end{equation}
where an exponential map$ \text{Exp}_{x'}$ is also applied afterwards to map the learned feature back to the manifold. Further, the authors also moved the non-linear activation function to the tangent space, since they found that the hyperbolic operation will collapse to the vanilla Euclidean GCN as the exponential map$ \text{Exp}_{x'}$ will be canceled by the logarithmic map at next layer. 

Work~\cite{peng2020mix} shares the same idea to construct spatial temporal graph convolutional networks~\cite{yan2018spatial} in hyperbolic space and applied this for dynamic graph sequences (human skeleton sequences in action recognition task). In addition, they further explored the projection dimension in the tangent space, instead of manually adjusting the projection dimension on the tangent spaces, they provided an efficient way to search the optimal dimension using neural architecture search~\cite{elsken2019neural}.

Different from previous methods, hyperbolic graph convolutional network~\cite{chami2019hyperbolic} decoupled the message passing procedure of GNN before they introduced their way to extend it to hyperbolic space.  Specifically, the operation of GNN is divided into three parts, including feature transform, neighborhood aggregation, and activating by the activation function. Formally, the operation of GNN is defined as
\begin{equation}
    \begin{split}
         h_i^{k} &=  \mathcal{W}^k h_i^{k} + b^k,\\
         h_i^{k} &=  h_i^{k} + \sum_{j \in \mathit{N}(i)}A_{ij} h_j^{k},\\
         h_i^{k+1} &= \sigma( h_i^{k})
    \end{split}
\end{equation}

After this decoupling, the author provided operations for corresponding parts in the hyperbolic space, which are 

\begin{equation}\label{eq:Hgnn}
    \begin{split}
         h_i^{k} &=  \mathcal{W}^k \otimes h_i^{k} \oplus b^k \\
         h_i^{k} &=   \mathcal{AGG} (h_i^{k}),\\
          h_i^{k+1} &= \text{Exp}_0 (\sigma(  \text{Log}_0 (h_i^{k}))
    \end{split}
\end{equation}

where the $\mathcal{W}^k \otimes h_i^{k}$ can be further computed by the exp and log maps. Let the intermediate output be $\hat{h_i^{k}}$, then $\hat{h_i^{k}}= \mathcal{W}^k \otimes h_i^{k} = \text{Exp}_0 ( \mathcal{W}^k \text{Log}_0 ( h_i^{k}) ).$ Then the bias is also added by using the parallel transport, which is $ h_i^{k} = \text{Exp}_{\hat{h_i^{k}}} (\mathcal{PT}_{0\rightarrow \hat{h_i^{k}}}(b^k))$. After this feature transformation, a neighborhood aggregation $ \mathcal{AGG}$ on the Lorentz model is introduced, which is

\begin{equation}\label{eq:Hgnn}
 \mathcal{AGG} (h_i^{k})=  \text{Exp}_{h_j^{k}} \left(\sum_{j \in \mathit{N}(i)}w_{ij} h_j^{k}\right).
\end{equation}
where the $w_{ij} = \textit{Softmax}_{j \in \mathcal{N}(i)} (MLP (\text{Log}_0 (h_i^{k}||h_j^{k})))$ works as the attention for the aggregation (Here, MLP is short for Euclidean Multi-layer Percerptron, and it is used to compute the attention based on the concatenated representation of the two nodes in the tangent space of point 0.) It is also worth mentioning that the information aggregation is performed at the tangent space of the center point of each neighbor set, which is different for different nodes and it is fair reasonable.

Work~\cite{bachmann2020constant} also presented a novel graph convolutional network in the Non-Euclidean space from the curvature perspective. The proposed network is mathematically grounded generalization of GCN to constant curvature spaces. Specifically, they extended the operations in gyrovector spaces to the space with constant positive curvature, which is the stereographic spherical projection models in their study. By this way, they provided a uniform GCN for spaces with different kinds of curvature ( 0, negative and positive), which is called the $\kappa$-GCN. Work~\cite{kochurov2020hyperbolic} further improved this method by providing a more reasonable definition of the gradient of curvature at zero, since the original one is incomplete. We will detail these methods in Sec.~\ref{sec:mixNN}.

\subsection{Hyperbolic Normalizing Flows}

Normalizing flows~\cite{rezende2015variational} involve learning a series of invertible transformations, which are used to transform a sample from a simple base distribution to a sample from a richer distribution. The models produce tractable distributions where both sampling and density evaluation can be efficient and exact.  For more details please refer to this survey paper~\cite{kobyzev2020normalizing}.

Mathematically, starting from a sample from a base distribution, $ z_0 \sim p(z)$, an invertible and a smooth mapping $f_{\theta}: \mathbb{R}^n \rightarrow \mathbb{R}^n$, the final sample from a flow is then given by 
a composition of functions, which is $z_j = f_j \o f_{j-1}\o...,\o f_1(z_0)$. The corresponding density can be determined simply by
\begin{equation}
    ln p_{\theta} (f_{\theta}(z_j)) = ln(p(z_0)) - \sum_{i=1}^{j}ln~det|\frac{\partial f_i}{\partial z_{j-1}}|.
\end{equation}
In practice computing the log determinant of the Jacobian is computationally expensive. Fortunately, appropriate choice of function $f$ can bring down the computation cost significantly. 

In fact, normalizing flows have already been extended to Riemannian manifolds (spherical spaces)~\cite{gemici2016normalizing,rezende2020normalizing}. Currently, we can also find normalizing flows being extended to Torus spaces~\cite{rezende2020normalizing}. In the work~\cite{brehmer2020flows}, they further extended it to a data manifold by proposing a new class of generative models, $\mathcal{M}$-flows, which simultaneously learn the data manifold as well as a tractable probability density on that manifold. Intended to benefit from the superior capability of hyperbolic space of modeling data with an underlying hierarchical structure, work~\cite{bose2020latent} presented a new normalizing flow in the hyperbolic space. They pointed out the problems of current Euclidean normalizing flows firstly, just like for other tasks, deep hierarchies in the data cannot be embedded in the Euclidean space without suffering from high distortion~\cite{sarkar2011low}. Secondly, sampling from densities defined on Euclidean space cannot make sure the generated points still lie on the underlying hyperbolic space. To address these issues, they proposed first elevated normalizing flows to hyperbolic space (leveraging Lorentz model) using coupling transforms defined on the tangent bundles. Then, they explicitly utilized the geometric structure of hyperbolic spaces and further introduced Wrapped Hyperboloid Coupling (WHC), which is a fully invertible and learnable transformation.

Normally, defining a normalizing flow in hyperbolic space by employing the tangent space at the origin is the easiest way.  Firstly, sample a point from a base distribution, which can be a Wrapped Normal~\cite{nagano2019wrapped}. Then transport the sample to the tangent space of the origin, using the logarithmic map. Next, apply any suitable Euclidean flows in the tangent space. Finally, project back the sample by using the exponential map. 

Based on this strategy, work~\cite{bose2020latent} provided a method called Tangent Coupling, which builds upon the real-valued non-volume preserving transformations (RealNVP flow)~\cite{dinh2016density} and introduces the efficient affine coupling layers. Specifically, this work follows normalizing flows designed with partially-ordered dependencies~\cite{dinh2016density}. They defined a class of invertible parametric hyperbolic functions $f_i^{\mathcal{TC}}: \mathbb{L}^k \rightarrow \mathbb{L}^k$. The coupling layer is implemented using a binary mask, and partitions the input x into two sets $x_1 = x_{1:d},x_2 = x_{d+1:n}$, For the first set $x_1$, its elements are transformed elementwise independently of other dimensions, while the transform of second set $x_2$ is based on the first one. Thus, the overall transformation of one layer is 
\begin{equation*}
    \hat{f}^{\mathcal{TC}} (\hat{x}) =  \left\{\begin{matrix}
 & \hat{z}_1 = \hat{x}_1 \\ 
 & \hat{z}_2 = \hat{x}_2 \odot \delta(s(\hat{x}_1))+ t(\hat{x}_1)
\end{matrix}\right.
\end{equation*}

\begin{equation}
    f^{\mathcal{TC}}(x) = \text{Exp}_0 (\hat{f}^{\mathcal{TC}} (\text{Log}_0 (x) )
\end{equation}
where $\odot$ and $\delta$ are pointwise multiplication and pointwise non-linearity, e.g., the exponential function, respectively.  $s$ and $t$ are map functions, which are implemented as linear neural layers and conduct the projection from $\mathcal{T}_o \mathbb{L}^d \rightarrow \mathcal{T}_o \mathbb{L}^{n-d} $. So the transformed result will be $\hat{z}$, which is the concatenation of resulted and mapped $\hat{z}_1$ and $\hat{z}_2$ on the manifold.  They also provided an efficient expression for the Jocobian determinant by using the chain rule and identity~\cite{bose2020latent}.

To make full use of the expression power of the manifold, they provided a new method, Wrapped Hyperboloid Coupling ($W\mathbb{H}C$), which improves the performance by using parallel transport, instead of only operating in the tangent space of origin. In particular, the $W\mathbb{H}C$ is defined as follows:
\begin{equation*}
    \hat{f}^{W\mathbb{H}C} (\hat{x}) =  \left\{\begin{matrix}
 & \hat{z}_1 = \hat{x}_1 \\ 
 & \hat{z}_2 = \text{Log}_0 \left( \text{Exp}_{\hat{x}_1} ( \mathcal{PT}_{0 \rightarrow t(\hat{x}_1)} (v)) \right),
\end{matrix}\right.
\end{equation*}
\begin{equation*}
    v = \hat{x}_2 \odot \delta(s(\hat{x}_1)),
\end{equation*}
\begin{equation}\label{eq:whc}
    f^{W\mathbb{H}C}(x) = \text{Exp}_0 (\hat{f}^{W\mathbb{H}C} (\text{Log}_0 (x) ),
\end{equation}

where $s: \mathcal{T}_o \mathbb{L}^d \rightarrow \mathcal{T}_o \mathbb{L}^{n-d}$ and $t: \mathcal{T}_o \mathbb{L}^d \rightarrow \mathcal{T}_o \mathbb{L}^{n}$ are map functions and also implemented as neural networks. However, the role of the function $t$ is changed. Here, since parallel transport is used in Eq.~(\ref{eq:whc}), $t$ can learn an optimal point on the manifold such that the tangent space transports from origin to this point. Likewise, they provided an efficient expression for the Jocobian determinant. In this way, they built two different kinds of normalizing flows in the Lorentz model. 





\subsection{Hyperbolic Variational Auto-Encoders}


Variational auto-encoders (VAEs)~\cite{kingma2014autoencoding,rezende2014stochastic} are a popular probabilistic generative model, composed of an auxiliary encoder that draws samples of latent code from the approximate posterior (conditional density), and a decoder generating observations $x \in \mathcal{X}$ from latent variables $z \in \mathcal{Z}$.  The objective of VAEs is to optimize the log-likelihood of the data, $log P_{\theta}(x) = log \int P_{\theta}(x,z) dz $, where $P_{\theta}(x,z)$ denotes a parameterized model of the joint distribution. However, marginalizing over the latent variables is generally intractable. Basically, the VAE networks leverages the approximate posterior to derive an evidence lower bound (ELBO) to the (intractable) marginal log-likelihood, which is
\begin{equation}\label{eq:elbo}
    \mathcal{L}(x;\theta,\phi) = \mathbb{E}_{z\sim q_{\phi}(z|x)}[log P_{\theta}(x|z)] - \mathbb{D}_{KL}(q_{\phi}(z|x)||p(z)),
\end{equation}
where the first conditional likelihood and the second KL terms respectively characterize reconstruction and generalization capabilities.  This ELBO can be further efficiently approximated by using the famous reparameterization trick~\cite{kingma2014autoencoding,rezende2014stochastic}, which is done by expressing a sample of $z\sim q_{\phi}(z|x)$ in Eq.~(\ref{eq:elbo}) using a reparameterization transformation and some random noise variables independent to the model $\theta$.

In addition, VAEs in non-Euclidean latent space have been recently introduced. For instance, Davidson \etal~ made use of hyperspherical geometry~\cite{s-vae18} and Falorsi~\etal endowing the latent space with a Lie group structure~\cite{falorsi2019reparameterizing}.

To this end, they proposed an efficient and reparametrizable sampling scheme, and calculated the probability density functions, for two canonical Gaussian generalisations defined on the Poincar\'e ball, namely the maximum-entropy and wrapped normal distributions. At the same time, they introduced a decoder architecture that explicitly takes into account the hyperbolic geometry.

The Gaussian distribution is a common choice of prior for vanilla VAE style models. One of the main challenges is how to generalize the latent distribution learning in the Euclidean space to the hyperbolic space.  As mentioned in~\cite{mathieu2019continuous,rezende2020normalizing}, there are mainly three ways to model the normal distributions in the hyperbolic space. 

Firstly, the Riemannian normal~\cite{pennec2006intrinsic} with Fr\'echet mean method~\cite{Frechetmean1948} $\mu$ and dispersion parameter $\sigma$. Sometimes, it is also referred as maximum entropy normal. In particular, the Riemannian normal distribution is defined as
\begin{equation}\label{eq:Rgaussion}
    \mathcal{N}_{\mathcal{M}}(z|\mu,\sigma^2) = \frac{1}{Z(\sigma)} \text{exp~}(\frac{-d_{\mathcal{M}}(\mu, z)}{2\sigma^2}) 
\end{equation}
where $d_{\mathcal{M}}$ is the induced distance~\cite{said2014new,mathieu2019continuous} and $Z(\sigma)$ is a normalization constant.

\begin{figure*}
    \centering
    \includegraphics[width= 0.8\textwidth]{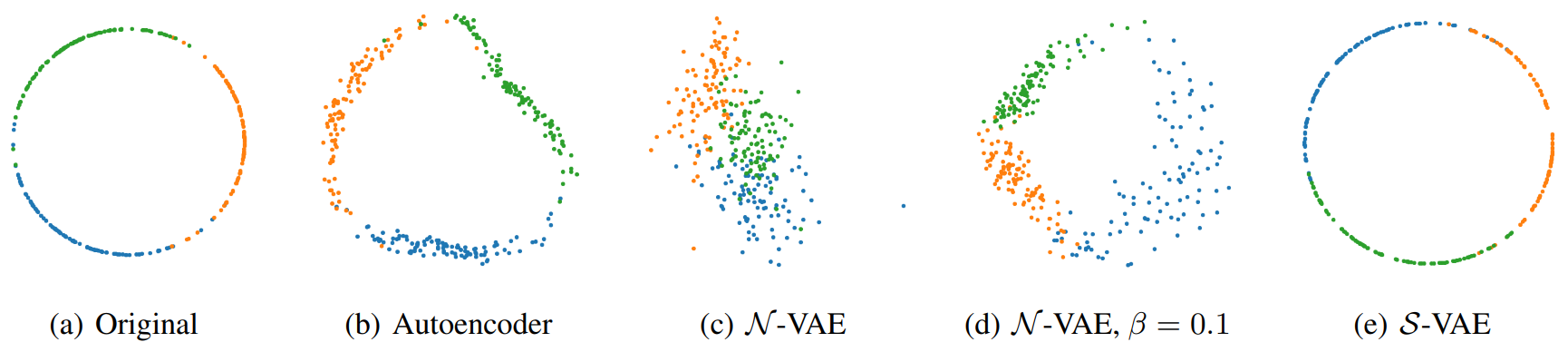}
    \caption{Plots of the original latent space (a) and learned latent space representations in different settings, where $\beta$ is a
re-scaling factor for weighting the KL divergence. $\mathcal{N}-$VAE represents vanilla VAE with Euclidean normal distribution. For more details please refer to~\cite{s-vae18}.}
    \label{fig:restrct-normal}
\end{figure*}

Secondly, the Restricted Normal. Restricting the sampled points from the normal distribution to sub-manifolds. It has also been treated as the maximum entropy distribution with respect to the ambient euclidean metric~\cite{mardia2014statistics}. For instance, in the work Hyperspherical variational auto-encoders~\cite{s-vae18}, the authors presented a novel VAE model, called $\mathcal{S}$-VAE, via von Mises-Fisher (vMF) distribution, in which the encoder is a homeomorphism and can provide an invertible and globally continuous mapping. In Fig.~\ref{fig:restrct-normal}, one can also see the visualization of the $\mathcal{S}$-VAE with this distribution compared to other autoencoder models. 

Thirdly, Wrapped Normal~\cite{falorsi2019reparameterizing,nagano2019wrapped,mathieu2019continuous}. This distribution is constructed by utilizing the exponential map of a Gaussian distribution on the tangent space centered at the mean value. Specifically, there are four steps to get a wrapped normal distribution. Let the the Wrapped Normal be $\mathcal{N}_{\mathcal{M}}(\mu, \sigma)$, which is constructed by First, sampling one point from the Euclidean normal distribution $\mathcal{N}(0, \sigma)$. Second, concatenate 0 as the zeroth coordinate of this point and transfer it to the tangent space of the origin. Third, parallel transports from the current tangent space to the tangent space at the point $\mu$. Finally, mapping the point from the tangent space to the manifold using the exponential map. In this way, a latent sample on the manifold is obtained. As mentioned in~\cite{dai2020apoVAE}, the Wrapped Normal has the following reparametrizable form
\begin{equation}\label{eq:wrapNorm}
    z = \text{Exp}_{\mu}(\mathcal{PT}_{0\rightarrow \mu}(v))), v \in \mathcal{N}(0, \sigma).
\end{equation}

As mentioned in~\cite{skopek2020mixed}, a downside of vMF is that it has only a single scalar covariance parameter, while other approaches can parametrize the covariance in different dimensions separately. The Riemannian Normal distributions could be computationally expensive for sampling if it is based on rejection sampling. On the contrary, Wrapped Normal distributions are very computationally efficient to sample.

In addition to the three methods mentioned above, there are also other ways to model the distribution in the manifold. For instance, the diffusion normal through the heat kernel~\cite{hsu2008brief,paeng2011brownian}. One can refer to work~\cite{mathieu2019continuous} or other related materials for more details. Based on the aforementioned generalization of the normal distribution in the non-Euclidean space, there are numerous works~\cite{ovinnikov2019poincar,grattarola2019adversarial,nagano2019wrapped,mathieu2019continuous,dai2020apoVAE} constructing VAE in the hyperbolic space, aiming at imposing structure information on the latent space.

Work~\cite{grattarola2019adversarial} proposed a general framework for embedding data on constant curvature manifolds (CCM). To this end, a CCM latent space within an adversarial auto-encoder framework is introduced. In this work, a prior distribution is constructed by using the tangent space with local Euclidean property. However, as mentioned by work~\cite{mathieu2019continuous}, the encoder and decoder are not designed to explicitly take into account the latent space geometry.

Work~\cite{ovinnikov2019poincar} proposed to endow a VAE latent space with the ability to model underlying structure via the Poincar\'e ball model. The author mentioned that the vanilla VAE posterior is parameterized as a unimodal distribution such that it is not able to allow the structure assumption. To deal with this problem, they provided a closed form definition of Gaussian distribution in the hyperbolic space, as well as the sampling rules for the prior and posterior distribution. Specifically, based on the maximum entropy generalization of Guassian distribution ~\cite{pennec2006intrinsic} (also the Riemannian normal mentioned above), they derived the normalization constant in Eq.(~\ref{eq:Rgaussion}) by decomposing it into radial and angular components. Based on the Wasserstein Autoencoder~\cite{tolstikhin2018wasserstein} framework, which is introduced to circumvent the high variance associated with the Monte-Carlo approximation, they built each layer using the hyperbolic feedforward layer provided by~\cite{ganea2018hyperbolic}. They also provided a generalization of the reparameterization trick by using the M\"{o}bius transformations. They further relaxed the constraint to the posterior by using the maximum mean discrepancy~\cite{borgwardt2006integrating} and the network is optimized by RSGD~\cite{bonnabel2013stochastic}. However, as pointed out by~\cite{mathieu2019continuous}, the authors had to choose a Wasserstein Auto-Encoder framework since they could not derive a closed-form solution of the ELBO’s entropy term. Besides, work~\cite{nagano2019wrapped} mentioned that the approximation of the likelihood and its gradient can be avoided.

Work~\cite{nagano2019wrapped}  provided a new normal distribution function in the hyperbolic space (Lorentz model), which is called Pseudo-Hyperbolic Gaussian, and it can be utilized to construct and learn a probabilistic model like VAE in this non-Euclidean space. The authors emphasise that this distribution is computed analytically and could be sampled efficiently. Pseudo-Hyperbolic Gaussian can be constructed with four steps, as mentioned above in the wrapped normal. The author further highlighted their contributions by deriving the density of Pseudo-Hyperbolic Gaussian distribution $\mathcal{G}(\mu,\Sigma)$ due to the exponential map and the parallel transport in the wrapped normal are differentiable.  

\begin{equation}
\begin{split}
    log~& p(z) = \\
    & log~ p(v) - log~\text{det }\left( \frac{\partial \text{Exp}_{\mu}(\mu)}{\partial \mu} \right) \text{det }\left( \frac{\partial \mathcal{PT}_{\mu_0 \rightarrow \mu(v)}(\mu)}{\partial v} \right),
\end{split}
\end{equation}
where $det$ denotes the determinant. Then they chose an orthonormal basis of tangent space $\mathcal{T}_{\mu}\mathbb{L}$ to evaluate the determinant of $\frac{\partial \text{Exp}_{\mu}(\mu)}{\partial \mu}$, which is derived as
\begin{equation}
    \text{det }\left( \frac{\partial \text{Exp}_{\mu}(\mu)}{\partial \mu}\right) = \left( \frac{\text{~sinh~}||\mu||_{\mathbb{L}}}{||\mu||_{\mathbb{L}}}\right)
\end{equation}


Since they provide a closed form of the density function, they could evaluate the ELBO exactly and there is no need to introduce the reparameterization trick in this hyperbolic VAE.

However, as pointed out by~\cite{mathieu2019continuous}, the neural layers are still the Euclidean ones, which did not take into account the hyperbolic geometry. Therefore, work~\cite{mathieu2019continuous} introduced a VAE that respects the geometry of the hyperbolic latent space. This is done by adding a generalization of the decision hyperplane in Euclidean space. Normally, the Euclidean linear affine transformation is $f(z)=sign(<a, z-p>)||a||d(z, H_{a,p})$, where $a$ is the coefficient, $p$ is the intercept (offset). The $H_{a,p}$ denotes a hyperplane passing through $p$ with $a$ as the normal direction, thus $d(z, H_{a,b})$ means the Euclidean distance of $z$ to the hyperplane. Analogues to the Euclidean linear function $f(z)$, they generalized it like
\begin{equation}
    f_{a,p}(z) = sign (<a, \text{Log}_p (z)>)||a||_p d^{\mathbb{B}}(z, H_{a,p}^{\mathbb{B}})
\end{equation}

\begin{figure}
    \centering
    \includegraphics[width = 0.45 \textwidth]{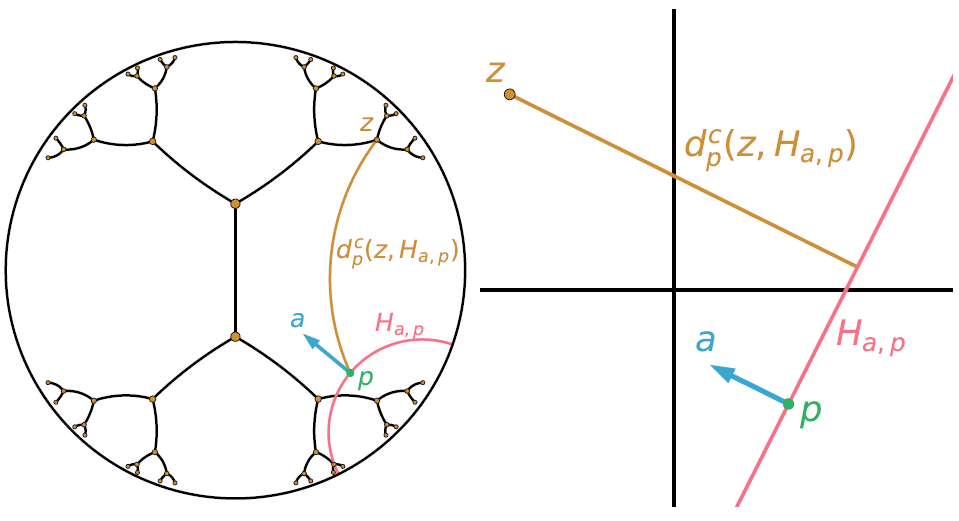}
    \caption{Hyperplane in Poincar\'e model from work~\cite{mathieu2019continuous}. Illustration of an orthogonal projection on a hyperplane in a Poincar\'e disc (Left) and an Euclidean plane (right).}
    \label{fig:Hyperplane}
\end{figure}

Inspired by the MLR in work~\cite{ganea2018hyperbolic}, the first layer of the decoder, which is called the gyroplane layer, as shown in Fig.~\ref{fig:Hyperplane}, is chosen to be a concatenation with a Poincar\'e operator $f$. Then it is then composed with a standard feed-forward neural network.

The gyroplane layer is then composed with a standard feed-forward neural network. For the encoder part, the author also changed the last layer by adding an exponential map for the Fr\'echet mean, and a softplus function for the positive defined $\Sigma$. 

Then, the ELBO of VAE is optimized via an unbiased Monte Carlo (MC) estimator with two main Gaussian generalisations, which are wrapped normal and Riemannian normal generalizations.  Through a hyperbolic polar change of coordinates, they provided an efficient and reparameterizable sampling schemes to calculate the probability density functions.

Compared to the above-mentioned method, the work~\cite{dai2020apoVAE} highlights an implicit posterior and data-driven prior. Work~\cite{dai2020apoVAE} proposed an Adversarial Poincar\'e Variational Autoencoder (APo-VAE). In particular, a wrapped normal distribution is used as the prior and the variational posterior for more expressive generalization. However, they replaced the tangent space sampling step in Eq.~(\ref{eq:wrapNorm}) with a more flexible implicit distribution from $\mathcal{N}(0, \textbf{I})$, inspired by work~\cite{fang2019implicit}. Then, a geometry-aware Poincar\'e decoder is constructed, which shares the same idea as it for the decoder in work~\cite{mathieu2019continuous}. 


ApoVAE further optimized the variational bound by adversarially training this model by exploiting the primal-dual formulation of Kullback-Leibler (KL) divergence based on the Fenchel duality~\cite{rockafellar1966extension}. And the training procedure is following the training scheme of coupled variational Bayes (CVB) from work~\cite{dai2018coupled} and implicit VAE~\cite{fang2019implicit}. Meanwhile, inspired by~\cite{tomczak2018vae}, they replaced the prior with a data-driven alternative to reduce the induced bias. 

In addition to all these methods discussed above, the work~\cite{skopek2020mixed} is also worth mentioning. This work presented an autoencoder in non-Euclidean space. However, instead of only using the hyperbolic space, they extended the VAE to multiple spaces, which is called mixed-curvature Variational Autoencoders. We will give details it in Sec.\ref{sec:mixNN}.




\subsection{Neural Networks with Mixed Geometries}\label{sec:mixNN}

\begin{figure}
    \centering
    \includegraphics[width = 0.5 \textwidth]{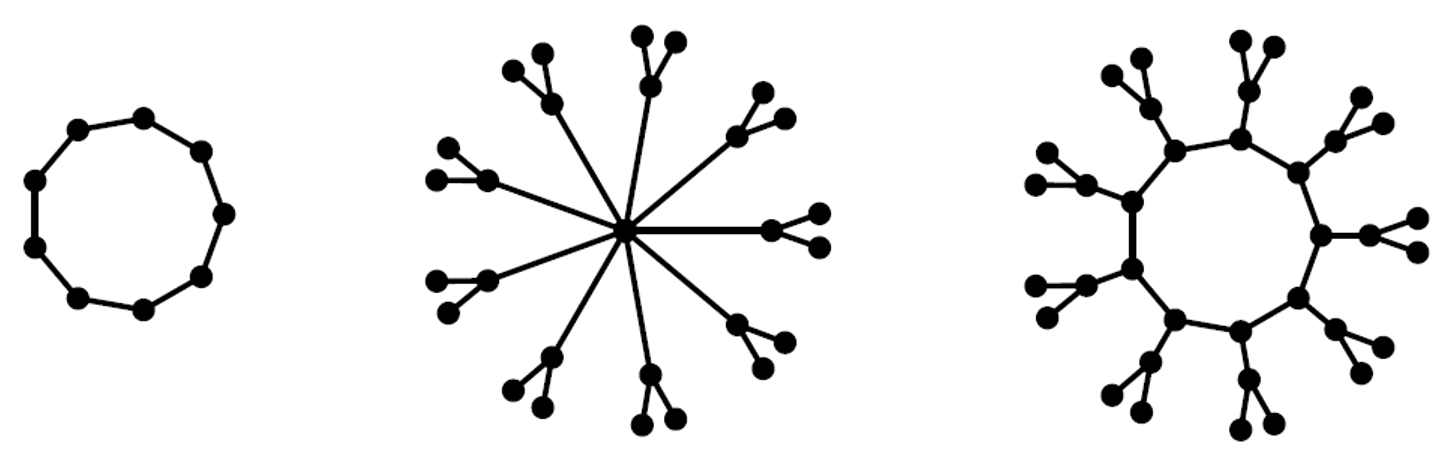}
    \caption{Illustration of data with different structures. Leftmost shows a data with cycle. Middle shows a tree-structured data and the rightmost is a combination.}
    \label{fig:structured-data}
\end{figure}

As mentioned by work~\cite{gu2018learning}, the quality of the learnt representations is determined by how well the geometry of the embedding space matches the underlying structure of the data. Hyperbolic space has already proved to be a better space to model tree-structured data. However, real-world data can be more complicated. For instance, in Fig.~\ref{fig:structured-data}, the leftmost cycle data can be better modeled by spherical space, the middle can be better modeled by the hyperbolic one. However, for the rightmost one, either of them is a good choice.  Instead, more flexible mixed spaces, providing a rich expressive representation for data, can be a choice for data involving multiple properties.

Currently, there is also an increasing momentum of learning representations via hybrid geometries, especially for data involved with a variety of types. Here, we summarize the literature mixing hyperbolic models with other geometries.

Work~\cite{grattarola2019adversarial} provided a unified autoencoder model for spaces with different kinds of curvatures, including spaces with positive, negative and also zero curvatures. However, this work did not take the geometry information into consideration when building the neural architecture, which is far away from the expectation.

Work~\cite{gu2018learning} pointed out that introducing a unique space may only be suitable for a certain type of structured data. However, most data is not structured so uniformly such that a mixed spaces would be helpful to the representation learning procedure.  In particular, they proposed to utilize a Riemannian product manifold of model spaces~\cite{lee2006riemannian}, including hyperbolic spaces with negative curvature, positively curved spherical space, and the flat Euclidean space. Given a sequence of smooth manifolds $\mathcal{M}_1,\mathcal{M}_2, \cdots ,\mathcal{M}_k$,  the product manifold is defined as a Cartesian product $\mathcal{M} = \mathcal{M}_1 \times \mathcal{M}_2,...,\times \mathcal{M}_k$. Here, each of the component spaces has a constant curvature and a point on this manifold $\mathcal{M}$ is $p=(p_1,...,p_k): p_i \in \mathcal{M}_i$ through all coordinates on each manifold $\mathcal{M}_i$.  As a result, there will be two challenges here, first, for each component space, how to determine the curvature. Second, the number of components of each space type and their dimensions, which is called signature by~\cite{gu2018learning}. To address these issues, the authors proposed to learn the curvature for each space. And for the choice of the number of spaces $k$ and the projection dimensions for each space, they introduced a heuristic estimator. Specifically, let $\mathbb{S}^d$, $\mathbb{H}^d$ and $\mathbb{E}^d$ be the spherical, hyperbolic, and Euclidean space with d dimensions, respectively. Then the product space $\mathcal{P}$ is represented by
\begin{equation}
    \mathcal{P} = \mathbb{S}^{s_1} \times\mathbb{S}^{s_2} \times \cdots \mathbb{S}^{s_m}\times \mathbb{H}^{h_1}\times \mathbb{H}^{s_2}\times \cdots \mathbb{S}^{h_n} \times \mathbb{E}^e,
\end{equation}
where $s_1,s_2,...s_m,h_1,h_2,...h_n,e$ are the dimensions for the corresponding spaces. Here, they only introduced one Euclidean space since the product of multiple Euclidean spaces is are isomorphic to the single space, while this is not the case for spherical and hyperbolic spaces~\cite{gu2018learning}. They introduced a graph embedding task with a loss objective 
to minimize the average distortion. This function can be further optimized by RSGD~\cite{bonnabel2013stochastic}.  Here, they highlighted the learning procedure by jointly optimizing the curvature of every non-Euclidean factor along with the embeddings. This is simply implemented by parameterizing the curvature in the distance metrics and then optimizing the non-Euclidean space in a unit space (by projecting the input into a unit space). For the signature, they matched the discrete notions of curvature on graphs with sectional curvature on manifolds and estimated it by triangle comparison theorem~\cite{toponogov2006differential}.

They also mentioned that the product manifold are at least as good as single spaces, while for more complicated structure, it can offer significant improvements.

The hyperbolic and spherical models chosen by work~\cite{gu2018learning} are Lorentz (hyperboloid) model and hypersphere model. However, as mentioned by~\cite{skopek2020mixed}, when the curvature $K$ goes to zero, both of these two spaces have an unsuitable property, named the non-convergence of the norm of points. For instance, the origin of Lorentz is $(1/K, 0,0,...,0)$, of which the norm is $\frac{1}{K} \rightarrow \infty$ when curvature $K$ goes to zero.

Work~\cite{tifrea2019poincarGlove} adapted the Glove~\cite{pennington2014glove} algorithm to learn unsupervised word embeddings in this type of Riemannian manifolds. To this end, they proposed to embed words in a Cartesian product of hyperbolic spaces which they theoretically connected to the Gaussian word embeddings and their Fisher geometry. 

Sharing the same idea, work~\cite{skopek2020mixed} provided an mixed-curvature variational autoencoders, devising a framework towards modeling probability distributions in products of spaces with constant curvature, utilizing latent space constructed by a product of constant curvature Riemannian manifolds. Like introduced in previous section, the wrapped normal distribution is used to change the parametrization of the mean and covariance in the VAE forward pass, and the choice of prior and posterior distributions. For the hyperbolic and spherical models, they chose the Poincar\'e model and projected sphere, utilizing classical stereographic projections~\cite{lee1997introduction}. They further proposed the fixed curvature and the learnable curvature VAEs. For the fixed one, they selected curvature values from the range [0.25, 1.0] with the corresponding radius in the range [1.0, 2.0]. And for the learnable curvature, they just simply introduced a parameter, while instead of directly learning the curvature, they learn the radius since they mentioned that learning curvature is badly conditioned. Another important thing is also to determine the the number of components (spaces) and also the dimension for each of them.  They proposed to partition the space into many 2-dimensional components (one component will have 3 dimensions if the space dimension is odd). Then, all of these small spaces were initialized and trained as Euclidean components. After training for a while, the components were split into three groups, including one hyperbolic component, one spherical, and the last remains Euclidean.  Finally, they allowed learning the curvatures of all non-Euclidean components. As also mentioned by the authors, the method is not completely general.

With similar motivations, work~\cite{bachmann2020constant} presented a constant curvature graph convolutional network, which is a mathematically grounded generalization of the traditional GCN framework.  Work~\cite{bachmann2020constant} provides a unified formalism for spaces with constant curvature. Based on this, they generalized GCN framework to non-Euclidean space and proposed a unified GCN, named $\kappa$-GCN. In particular, the authors unified all curvature spaces (positive, negative, and zero) via a stereographic projection~\cite{barth1937stereographic}, which is a mapping (function) that projects a sphere/hyperbolic manifold onto a plane. With this $\kappa$-stereographic model, they extended gyrovector spaces to positive curvature. Therefore, the exponential maps for a space with curvature $\kappa$ are described as
\begin{equation}
    \text{Exp}_{x}^{\kappa}(v) = x \oplus_{\kappa}\left( tan_{\kappa} \left( ||v||/2 \right) \frac{v}{||v||}\right),
\end{equation}
where the $tan_{\kappa}$ is defined as
\begin{equation}\label{eq:kap}
   tan_{\kappa}(x) = \left\{\begin{matrix}
\kappa^{\frac{-1}{2}} tan(x\kappa^{\frac{1}{2}}) & \kappa > 0\\ 
x & \kappa = 0\\ 
|\kappa|^{\frac{-1}{2}} tan(x|\kappa|^{\frac{1}{2}}) &\kappa < 0.
\end{matrix}\right.
\end{equation}

Likewise, we can also extend it to logarithmic maps, parallel transport and distance. With this method, they derived a unified graph convolutional network, $\kappa$-GCN, which reduces to commonly used GCN when $\kappa \rightarrow 0$. 

However, the work~\cite{kochurov2020hyperbolic} found that at the point of zero curvature, the gradient is not well defined. Therefore, based on work~\cite{bachmann2020constant}, they provided a new model, $\kappa$-Stereographic to correct the gradient for curvature $\kappa$. As the authors mentioned, by taking left and right Taylor expansions, the $tan_{\kappa}$ in Eq.~(\ref{eq:kap}) is re-defined as
\begin{equation}
   tan_{\kappa}(x) = \left\{\begin{matrix}
\kappa^{\frac{-1}{2}} tan(x\kappa^{\frac{1}{2}}) & \kappa > 0\\ 
x+\frac{1}{3}\kappa x^{3} +\frac{2}{15}\kappa^{2}x^5 & \kappa = 0\\ 
|\kappa|^{\frac{-1}{2}} tan(x|\kappa|^{\frac{1}{2}}) &\kappa < 0.
\end{matrix}\right.
\end{equation}

The authors also provided good examples to study various optimization techniques in non-Euclidean space, which is very valuable for this community. Specifically, the optimization step changes the geometry of space and the parameter constraints are also dependent on the curvature parameter, thus curvature optimization in hyperbolic space is challenging. As summarized by~\cite{kochurov2020hyperbolic}, there are mainly three ways to conduct curvature optimization, including alternating optimization, joint optimization, and tangent space optimization. 

Work~\cite{law2020ultrahyperbolic} also noticed that the hierarchical datasets may also contain cycles, which cannot be represented using trees. Therefore, they provided a model using the pseudo-Riemannian manifold (also called semi-Riemannian~\cite{o1983semi}) with constant nonzero curvature, which is a generalization of hyperbolic and spherical geometries where the non-degenerate metric tensor is not positive definite. The comparison results on a social network graph even show a superiority to the hyperbolic space.

Previous methods with mixed geometries are mainly focused on spaces with different curvatures and their combination. Very limited works study the interaction between them. Work~\cite{peng2020mix} mix the dimensions of different manifolds and provided an efficient interaction way to search the optimal projection dimensions. However, they are in the same hyperbolic spaces. Work~\cite{zhu2020graph} developed more advanced architectures by the interaction of Euclidean or hyperbolic spaces, learning the representation of graph. Specifically, this geometry interaction learning method has two branches, e.g., Euclidean network and hyperbolic (Poincar\'e model) network, in which they propagated neighbor information simultaneously via a distance-aware attention mechanism. Then a dual feature interaction is introduced to enhance both features by exponential and logarithmic functions. Experiments prove the efficacy of the geometry interactions.






\section{Applications and performance}\label{Sec6}

Latent hierarchical structures are characteristic for many complex symbolic datasets, from various research fields. Therefore, we will introduce the applications for the data with hierarchical structures, including learning embeddings of graphs, NLP, tree-like properties. Then, we also notice that there is an increasing potential to introduce hyperbolic neural network to regular data like natural images. Therefore, we introduce how hyperbolic networks are adapted to computer vision tasks without clear structures.

\begin{table*}[t]
    \centering
    \caption{\textbf{Summary of the advanced machine learning methods in hyperbolic space.} Here, G. means the type of \textbf{G}eometry. Its value 'mixed' means the methods combine more than one different geometries to build the model.}
    \begin{tabular}{lcccccc}
    \toprule
    Method & Year & Architecture & Tasks & G. & Institution & Source\\
    \hline
    PEmbedding~\cite{nickel2017poincare}     & 2017 & Embedding & NLP and Graph & $\mathbb{B}$ & Facebook & NeurIPS \\
    \hline
      HyperGraph~\cite{chamberlain2017neural} & 2017& Embedding & Network Vertex Classification & $\mathbb{B}$ & ICL & MLG W \\
    \midrule
    \midrule
       TextHyper~\cite{dhingra2018embedding}  & 2018 & AE(GRUs) & Text Embedding & $\mathbb{B}$ & Google & NAACL W  \\
    \hline
      h-MDS~\cite{de2018representation}   & 2018 & Embedding & Tree and tree-like data modeling & $\mathbb{B}$ & Stanford & PMLR  \\
         
    \hline
      HyperQA~\cite{tay2018hyperbolic}   & 2018 & Encoder-Decoder & Neural Question Answering &$\mathbb{B}$ & NTU & WSDM \\
    \hline
        HyperCone~\cite{ganea2018hypercone} & 2018 & Embedding & NLP and Graph & $\mathbb{B}$ & ETH Z\"{u}rich & ICML \\
    \hline
         HNN~\cite{ganea2018hyperbolic}& 2018 & RNN & NLP(textual entailment and noisy prefixes) & $\mathbb{B}$& ETH Z\"{u}rich & NeurIPS \\
    
    \hline
         Lorentz~\cite{nickel2018learning}& 2018& Embedding & Taxonomies Embedding, Graph and Historical Linguistics & $\mathbb{L}$& Facebook & ICML  \\
    \hline
         HyperBPR~\cite{vinh2018hyperbolic}& 2018 & BPR~\cite{rendle2012bpr} & Recommender Systems & $\mathbb{B}$& NTU & AAAI \\
    \hline
         ProductM~\cite{gu2018learning} & 2018  & Embedding & Tree(with Cycle)& Mixed & Stanford & ICLR \\
    \midrule
    \midrule
         HAN~\cite{gulcehre2018hyperbolic} & 2019 & Attention Module& Graph, Machine translation and Relational Modeling& $\mathbb{L}$($\mathbb{K}$)& Deepmind & ICLR \\
    \hline
        HGCN~\cite{chami2019hyperbolic} &2019 & GNN & Graph & $\mathbb{L}$& Stanford & NeurIPS \\
    \hline
         PGlove~\cite{tifrea2019poincarGlove}& 2019 & Glove~\cite{pennington2014glove} & Word Embedding & Mixed &ETH Z\"{u}rich & ICLR \\
    \hline
        HGNN~\cite{liu2019hyperbolic} & 2019& GNN & Graph & $\mathbb{B}$,$\mathbb{L}$& Facebook& NeurIPS  \\
    \hline
        Tiling~\cite{yu2019numerically}&2019 & Tiling& NLP and Compressing & $\mathbb{L}$& Cornell & NeurIPS \\
    \hline
       PTaxo~\cite{aly2019every} &2019 & Embedding& Taxonomy Induction & $\mathbb{B}$& U. Hamburg& ACL \\
    \hline
        H-SVM~\cite{cho2019large} &2019 & SVM & NLP(Word E) Graph (Node C) & $\mathbb{B}$& MIT & AISTATS \\
    \hline
       H-Recom~\cite{chamberlain2019scalable}  & 2019 & BPR & Recommender Systems & $\mathbb{L}$& ASOS & CoRR \\
    \hline
       MuRP~\cite{balazevic2019multi}  & 2019 & Bilinear & knowledge Graph & $\mathbb{B}$& U. Edinburgh & NeurIPS  \\
    \hline
       RAO~\cite{bcigneul2019riemannian}  & 2019 & Optimizor & NLP & $\mathbb{B}$& ETH Z\"{u}rich & ICLR \\
    \hline
      WrapN~\cite{nagano2019wrapped}   & 2019 & VAE & NLP,MNIST and Atari trajectories & $\mathbb{B}$& U.Tokyo & ICML  \\
    \hline
       LDistance~\cite{law2019lorentzian} & 2019 & Embedding & NLP & $\mathbb{B}$& U. Toronto& ICML \\
    \hline
      PVAE~\cite{mathieu2019continuous}   &2019 & VAE & NLP, and MNIST & $\mathbb{B}$& Oxford & NeurIPS \\
    \hline 
      CCM-AAE~\cite{grattarola2019adversarial} & 2019 & AE & MNIST C, Graph& Mixed & U.Lugano & ASOC  \\
    \hline
       HWAE~\cite{ovinnikov2019poincar}  & 2019 & VAE & G MNIST, Graph and Tree & $\mathbb{B}$& ETH Z\"{u}rich & - \\
    \hline
       gHHC~\cite{monath2019gradient}  & 2019 & Clustering & Clustering ImagNet, Multi-Task & $\mathbb{B}$& U. Mass & KDD \\
    \midrule 
    \midrule
       HGAT~\cite{zhang2020hyperbolic}  & 2020 & GNN & Graph & $\mathbb{B}$& BUPT & AAAI \\
     \hline
         H-STGCN~\cite{peng2020mix} & 2020 & GNN & Skeleton Action & $\mathbb{B}$& U. Oulu & ACM MM \\
    \hline
      HyperKG~\cite{kolyvakis2019hyperkg} & 2020 & Translational & Knowledge Graph& $\mathbb{B}$& EPFL& ESWC \\
    \hline
        HyperML~\cite{tran2020hyperml} & 2020 & Metric Learning & Recommender Systems& $\mathbb{B}$ & NTU & WSDM\\
    \hline
       LorentzFM~\cite{xu2020learning}  & 2020& Triangle Inequality & Recommender Systems & $\mathbb{B}$& eBay & AAAI  \\
    \hline
      APo-VAE~\cite{dai2020apoVAE} & 2020 & VAE & NLP (dialog-response generation)& $\mathbb{B}$& Duke& - \\
    \hline
       H-Image~\cite{khrulkov2020hyperbolic}  & 2020 & Embedding & Image c, few-shot & $\mathbb{B}$& SIST & CVPR \\
    \hline
       HyperText~\cite{zhu2020hypertext}  & 2020 & RNN & NLP(text classification)& $\mathbb{B}$& Huawei & EMNLP \\
    \hline
       ATTH~\cite{chami2020low}  & 2020 & RNN &  Knowledge Graph & $\mathbb{B}$& Stanford & ACL \\
    \hline
       $\kappa$-GCN~\cite{bachmann2020constant}  & 2020 & GNN & Graph & Mixed & ETH Z\"{u}rich & ICML \\
    \hline
        FMean~\cite{lou2020differentiating} & 2020 & GNN & Graph & $\mathbb{B}$& Cornell & ICML \\
    \hline
        H-NormF~\cite{bose2020latent} & 2020& Norm. Flow& Graph & $\mathbb{B}$& McGill& ICML \\
    \hline 
        $\kappa$-Stereographic~\cite{kochurov2020hyperbolic} & 2020 & GNN & Graph & Mixed & SIST & ICML \\
    \hline 
        L-Group~\cite{bogatskiy2020lorentz} & 2020 & Group & jet physics & $\mathbb{L}$ &U. Chicago & ICML \\
    \hline
        MVAE~\cite{skopek2020mixed} &2020 & VAE & Image reconstruction and Tree systhsis & Mixed & ETH Z\"{u}rich & ICLR \\
    \hline
        HypHC~\cite{chami2020trees} & 2020 & Clustering & Clustering(e.g., CIFAR-100)& $\mathbb{B}$& Stanford & NeurIPS \\
     \hline
        R-NormF~\cite{mathieu2020riemannian} &2020 & Norm. Flow& Earth sciences& $\mathbb{B}$& Oxford & NeurIPS \\
    \hline
         RH-SVM~\cite{weber2020robust}& 2020 & SVM & ImageNet(Pick 2classes) & $\mathbb{L}$& Princeton & NeurIPS \\
    \hline
       TREEREP~\cite{sonthalia2020tree}  &2020 & Embedding & Tree & $\mathbb{R}$ &U. Michigan & NeurIPS \\
    \hline
         UltraH~\cite{law2020ultrahyperbolic} & 2020 & Embedding & Graph & Mixed& NVIDIA & NeurIPS \\
    \hline
       GIL~\cite{zhu2020graph}  & 2020 & GNN & Graph & $\mathbb{B}$& CAS & NeurIPS \\
    \midrule
    \midrule
         HNN++~\cite{shimizu2021hyperbolic} & 2021 & CNN,Transformer & NLP, Clustering, Machine Translation& $\mathbb{B}$& U. Tokyo& ICLR \\
    \bottomrule
    \end{tabular}
    \label{tab:summaryNN}
\end{table*}

\subsection{NLP with Hyperbolic Embeddings}

Natural language often conceives a latent hierarchical structure, e.g., linguistic ontology. Likewise, a more generic short phrase can have many plausible continuations and is therefore semantically related to a multitude of long phrases that are not necessarily closely related (in the semantic sense). Based on this, it is natural to transfer natural language processing (NLP) from Euclidean space to hyperbolic space due to its innate suitability to embed hierarchies with low distortion~\cite{sarkar2011low}. As mentioned by work~\cite{dai2020apoVAE}, another advantage of modeling in hyperbolic space is that the latent representation gives us more control of the sentences we want to generate, which means  an increase of sentence complexity and diversity can be achieved along some trajectory from a root to its children. As a result, numerous works are presented to take advantage of hyperbolic neural networks for various NLP tasks. As summarized in Table~\ref{tab:summaryNN}, there are more than one third hyperbolic methods being presented to deal with NLP tasks, of which specific tasks include text classification~\cite{zhu2020hypertext}, entity typing~\cite{lopez2019fine,lopez2020fully}, taxonomy induction~\cite{aly2019every}, taxonomies embedding~\cite{nickel2018learning}, word embeddings~\cite{tifrea2019poincarGlove,dhingra2018embedding}, Lexical Entailment~\cite{nickel2017poincare} and text generation~\cite{dai2020apoVAE}. 

Among them, the WordNet~\cite{miller1998wordnet} dataset is very commonly used for evaluation. This dataset is a large lexical database which, amongst other relations, provides hypernymy (is-a(v,m)) relations, such as is-a(dog,animal). In this dataset, the noun and verb hierarchy of WordNet. The transitive closure of the WordNet noun hierarchy consists of 82,115 nouns and 743,241 hypernymy relations. For WordNet verb, there are 13,542 verbs and 35,079 hypernymy relations. 

In the task of \textbf{sentence entailment classification}, works~\cite{ganea2018hyperbolic} and ~\cite{shimizu2021hyperbolic} evaluated the proposed MLRs in the sentence entailment classification task. The results confirm the tendency of the hyperbolic MLRs to outperform the Euclidean version in all settings. At the same time, the hyperbolic MLR from work~\cite{shimizu2021hyperbolic} shows more stable training, relatively narrower confidence intervals, and at least comparable performance with only half of the parameters compared to the MLR in~\cite{ganea2018hyperbolic}.

In the \textbf{neural machine translation} task, the Transformer~\cite{vaswani2017attention} is a recently introduced  state-of-the-art model. Based on the hyperbolic attention mechanism, work~\cite{gulcehre2018hyperbolic} provided a hyperbolic version of Transformer, which one can observe the improvements over the original Euclidean Transformer from experiments. The improvements are more significant when the model capacity is restricted. Work~\cite{shimizu2021hyperbolic} also evaluates their Poincar\'e CNNs in the machine translation of WMT’17 English-German~\cite{ondrej2017findings}. Comparison results demonstrate the large improvements compared to the usual Euclidean models in fewer dimensions. However, this superiority disappears when further increasing the embedding dimension.

In the task of \textbf{text classification}, HyperText~\cite{zhu2020hypertext} performs significantly better than FastText~\cite{joulin2017bag}, which is the state-of-the-art text classifier based on shallow neural networks in Euclidean space, on datasets with more label categories. Besides, HyperText with 50-dimension achieves better performance to FastText with 300-dimension, which proves the hyperbolic space is a better choice for this task. Also, works~\cite{zhu2020hypertext} and~\cite{micic2018hyperbolic} benefits from the hyperbolic methods in the Chinese text analysis tasks. 

In the \textbf{text generation} task, HyperQA from work~\cite{tay2018hyperbolic} is the first to model QA pairs in hyperbolic space. HyperQA is an extremely fast and parameter efficient model that achieves very competitive results on multiple QA benchmarks, especially when compared to Euclidean methods at that time.  ApoVAE~\cite{dai2020apoVAE} also deals with the dialog-response generation problem. It optimized the variational bound by adversarially training and exploited the primal-dual formulation of KL divergence based on the Fenchel duality ~\cite{rockafellar1966extension}. 

In the task of \textbf{word embeddings}, work~\cite{nagano2019wrapped} proposed a Gaussian-like distribution in hyperbolic space, which is called pseudo-hyperbolic Gaussian. Based on this a hyperbolic VAE is presented to deal with the word embeddings. Work~\cite{tifrea2019poincarGlove} adapted the Glove~\cite{pennington2014glove} algorithm to learn unsupervised word embeddings in this type of Riemannian manifolds. To this end, they proposed to embed words in a Cartesian product of hyperbolic spaces which they theoretically connected to the Gaussian word embeddings and their Fisher geometry. We can see some notable founds in the experiments of Hypernymy~\cite{vulic2017hyperlex} evaluation task.  Based on their method, the fully unsupervised model can almost outperform all supervised Euclidean counterparts. Once trained with a small amount of weakly supervision for the hypernymy score, they can obtain significant improvements and this result is much better that the models in Euclidean space.

In the \textbf{ taxonomy embedding} task, work~\cite{nickel2017poincare} is one pioneer research piece that can learn embeddings in hyperbolic space (Poincar\'e model). They performed experiments on the tasks of embedding of taxonomies. To evaluate the ability of the embeddings to infer hierarchical relationships \textbf{without supervision}, they trained on data where the hierarchy of objects is not explicitly encoded. A significant improvement was witnessed in the Taxonomy Embedding task. Work~\cite{dhingra2018embedding} proposed a re-parametrization of Poincar\'e embeddings that removes the need for the projection step and allows the use of any of the popular optimization techniques in deep learning, such as Adam.  Work~\cite{nickel2018learning} pointed out that the Lorentz model is substantially more efficient than in the Poincar\'e model. Therefore, they proposed to learn a continuous representation using Lorentz model, which further improves the performance in this task. Work~\cite{law2019lorentzian} mentioned that Poincar\'e distance is numerically unstable and explained why the squared Lorentzian distance is a better choice than the Poincar\'e metric. Based on this, they learnt a closed-form squared Lorentzian distance and thus further improve the performance on the Taxonomy Embedding task. Very recently, work~\cite{shimizu2021hyperbolic} shows the superior parameter efficiency of their methods compared to conventional hyperbolic components, and the stability and outperform over their Euclidean counterparts. They evaluated their Poincar\'e MLR on the WordNet with the one proposed in ~\cite{ganea2018hyperbolic}, as well as the counterpart in Euclidean space. As mentioned before, work~\cite{yu2019numerically} constructs the hyperbolic model for dealing with the numerical instability of the previous hyperbolic networks. Based on the Lorentz model, they provide a very efficient model to learn the embedding. For instance, they can even compress the embeddings down to 2\% of a Poincar\'e embedding of the onWordNet Nouns.

In the task of \textbf{taxonomy induction}, which is aiming at creating a semantic hierarchy of entities by using hyponym-hypernym relations, traditional approaches like extraction of lexicalsyntactic patterns~\cite{hearst1992automatic} and co-occurrence information~\cite{grefenstette-2015-inriasac} are still dominating in the state-of-the-art taxonomy learning. Currently, work \cite{aly2019every} proposed to employ Poincar\'e embeddings in this task. Specifically, they first created a domain-specific Poincar\'e embedding. Based on this, they identified and relocated the outliers, as well as attached the unconnected terms.

\textbf{Textual entailment}, which is also called natural language inference, is a binary classification task to predict whether the second sentence (hypothesis) can be inferred from the first one (premise). Dataset SNLI~\cite{bowman2015large} is one typical dataset for this task, which consists of 570K training, 10K validation and 10K test sentence pairs. In the textual entailment task, HNN~\cite{ganea2018hyperbolic} embedded two sentences using two distinct hyperbolic RNNs. With the corresponding distances, the sentence embeddings are then fed into a feedforward network and predicted with an MLR. Interestingly, the results shows that the fully Euclidean baselines might even have an advantage over hyperbolic models. On top of pre-trained Poincar\'e embeddings~\cite{nickel2017poincare}, they conducted experiments on the WordNet noun hierarchy to evaluate the hyperbolic MLR. On this subtree classification task, hyperbolic MLR displays a clear advantage to the Euclidean counterpart. Despite work HNN~\cite{ganea2018hyperbolic} is one big step for generalizing deep learning components to hyperbolic space, in what case hyperbolic is more suitable is not clear enough.


Entity typing is a multi-class multi-label classification problem, in which, given a context sentence containing an entity mention, the goal is to predict the correct type labels that describe the mention. Usually, the types come from a predefined type inventory. Lopez et al.~\cite{lopez2020fully} provided the first fully hyperbolic neural model for hierarchical multi-class classification, and apply it in Entity typing. Based on hyperbolic neural network~\cite{ganea2018hyperbolic}, they built an attention-based mention and context encoder to extract feature representations, followed by a hyperbolic multinomial logistic regression (MLR) to perform the predictions. They adapted the MLR such that several classes can be selected. They showed how the hyperbolic definition of the classification hyperplanes benefits the task, given the hierarchical structure of the type inventory.

\subsection{Hyperbolic for Graph applications}

Graph data and its applications increasingly attract attention since graphs can be found everywhere in the real-life world and they can be utilized to model more complex underlying relationship than regular data. At the same time, hierarchies are ubiquitous in graph data. There are a great deal of methods presented to deal with real-world problems with graph data, including node classification~\cite{chami2019hyperbolic},  graph classification~\cite{peng2020mix,liu2019hyperbolic}, link prediction~\cite{chami2019hyperbolic}, and graph embedding~\cite{bachmann2020constant}. For instance, in the real-world dataset Amazon Photo~\cite{shchur2018pitfalls}, a node represents a product and an edge indicates that two goods are frequently bought together. All nodes in these datasets correspond to a label and a bag-of-words representation. Multiple graph tasks can be conducted on this dataset.

There are numerous works leveraging deep hyperbolic neural networks dealing with the graph tasks. Here, we only concentrate on the models generalized from the GNN architecture, although there are many other existing hyperbolic embedding methods, like PVAE~\cite{mathieu2019continuous}, HAT~\cite{gulcehre2018hyperbolic}, which also consider the modeling of graph (mainly for natural language or networks).

Work~\cite{chamberlain2017neural} is one of the pioneering works to deal with this topic. It introduces a new concept of neural embeddings in hyperbolic space.  They evaluated their method on five publicly available network datasets for the problem of vertex attribution. The results shows that it significantly outperforms Euclidean deepwalk~\cite{perozzi2014deepwalk} embeddings. Hyperbolic graph neural network (HGNN) in work~\cite{liu2019hyperbolic} and hyperbolic graph convolutional network (HGCN) in the work~\cite{chami2019hyperbolic} are almost proposed at the same time. They are the pioneer works for generalizing graph neural networks to hyperbolic space, using the tangent space. The HGCN is built on the Lorentz model. The authors tested HGCN on both node classification and link prediction tasks. Experiments show that for a dataset with low $\delta$-hyperbolicity, the HGCN can get a performance much better than models built in Euclidean space.  The HGNN is built on both the Poincar\'e model and Lorentz model. The authors dealt with both graph classification tasks, in which the graphs are synthetically generated with three distinct graph generation algorithms. The results show that when the embedding dimension is smaller than 20, the HGNN shows a significant superiority. We can also see that for most cases, HGNN on the Lorentz model performs better than that on the Poincar\'e model. While the authors increased the dimension, this advantages disappeared and in the task of graph classification, when the dimension is larger than 256, the HGNN performs worse that its Euclidean counterpart. Since the networks are built in the tangent space, the authors also explore the effectiveness of the activation function. They concluded that the activation function should perform on the output of the exponential map, otherwise multiple layers of HGNN would collapse to a vanilla Euclidean GCN. However, they do not evaluate whether the features are still on the manifold after performing the activation function on the manifold. 

Currently, work~\cite{bachmann2020constant} derives a general version of GCN with constant cavature, $\kappa$-GCN. In this case, the vanilla Euclidean GCN is just a special case of $\kappa$-GCN, with $\kappa = 0$. It is worth mentioning that this $\kappa$-GCN is not built on the tangent space. This model can significantly minimize the graph embedding distortion and get a superior performance on the node classification tasks. 



Most of previous mentioned tasks are focused on static graphs, while work~\cite{peng2020mix} considered the graph classification task with a dynamic graph input. They constructed and searched for the optimal ST-GCN in the Poincar\'e model. Another very interesting work~\cite{kochurov2020hyperbolic} is trying to answer the question that are hyperbolic representations of graphs created equal? The author doubted whether the hyperbolic spaces always give a superior performance when compared to the Euclidean counterparts.  To this end, work~\cite{kochurov2020hyperbolic} provided a profound analysis by conducting extensive experiments on four different kinds of graph tasks, including node classification, link prediction,  graph classification, and graph embedding. The experimental results suggest that the non-Euclidean space are not always a better choice that the Euclidean counterpart. As summarized by work~\cite{kochurov2020hyperbolic}, in the task that labels depend only on the local neighborhood of the node, hyperbolic models may be inferior to their Euclidean counterparts. However, many other factors, like the the way to build the manifold and the corresponding optimization strategy, may also lead to this result. Therefore, more explorations are needed to draw this conclusion.

\subsection{Hyperbolic Space for Recommender Systems}

Recommender systems are a pervasive technology, providing a major source of revenue and user satisfaction in online digital businesses~\cite{gomez2015netflix}. Recommender systems and ads click-trough rate (CTR) prediction both have a profound impact on web-scale big data in e-commerce, where due to large catalogue sizes, customers are often unaware of the full extent of available products~\cite{chamberlain2019scalable}. One of the most important factor to the success of a recommender system is the accurate representation of user preferences and item characteristics. To this end, matrix factorization~\cite{hu2008collaborative} is one of the most common approaches for this task. To approximate the original interaction matrix (huge),  a factorization is performed such that the resulted two low-rank matrices, representing users and items respectively, proving a much efficient way to compute the interaction, in the Euclidean vector space. The user-item interaction matrix is then treated as the adjacency matrix of an undirected, bipartite graphs, which exhibit clustering produced by similar users interacting with similar items, thus modeled as the complex networks~\cite{newman2003structure}. 

As mentioned in the page titled hyperbolic Geometry of Complex Networks ~\cite{krioukov2010hyperbolic}, hyperbolic geometry naturally emerges from network heterogeneity in the same way that network heterogeneity emerges from hyperbolic geometry. Therefore, given the complex nature of these networks, a hyperbolic space is better suited to embed them than its Euclidean counterpart.

Based on the above observation, work~\cite{chamberlain2019scalable} that embeds bipartite user-item graphs in hyperbolic space can provide such a recommender system. This recommendation algorithm learns to rank loss that represents users and items, uses of hyperbolic representations for neural
recommender systems through an analogy with complex networks. This algorithm shows a clear advantage when compared with the system in the Euclidean space. And the system based on this algorithm is scaled to millions of users.

Although this system shows obvious superiority, work~\cite{tran2020hyperml} pointed out that work~\cite{chamberlain2019scalable} does not learn the embeddings with metric learning manner. Therefore, work~\cite{tran2020hyperml}  explored the connections between metric learning in hyperbolic space and collaborative filtering and proposed a highly effective model for recommender systems. They authors constructed an input triplet tuple with user, item liked by user and item unliked by user. Then they learnt the user-item joint metric in the hyperbolic space (Poincar\'e model). At the same time, they also introduced so-called local and global factor to better embed user-item pairs to hyperbolic space and at the same time preserve good structure quality for metric learning.

Work~\cite{xu2020learning} pointed out that in the factorization machine model~\cite{rendle2010factorization}, the naive inner product is not expressive enough for spurious or implicit feature interactions. Therefore, higher-order factorization machine~\cite{blondel2016higher} is proposed to learn higher-order feature interactions efficiently. As suggested by collaborative metric learning~\cite{hsieh2017collaborative}, learning the distance instead of inner product has advantages to learn a fine-grained embedding space which could capture the representation for item-user interactions, item-item and user-user distances at the same time.  Thus, the triangle inequality is preferred than the inner product. Inspired by this, work~\cite{xu2020learning} proposed a model named Lorentzian Factorization Machine (LorentzFM), which learnt feature interactions with a score function measuring the validity of triangle inequalities. The authors argued that the feature interaction between two points can be learnt by the \textbf{sign} of the triangle inequality for Lorentz distance, rather than using the distance itself. Based on this, they presented the model in the hyperbolic space.


\subsection{Knowledge Graph completion}

Knowledge graph is a multi-relational graph representation of a collection of facts $\mathcal{F}$, formed in a set of triplet $(e_s,r,e_o)$, where $e_s$ is the subject entity, $e_o$ is the object entity and r is a binary relation (typed directed edges) between them. $(e_s,r,e_o) \in \mathcal{F}$ denotes subject entity $e_s$ is related to object entity $e_o$ by the relation r. As mentioned by work~\cite{balazevic2019multi}, knowledge graphs often exhibit multiple hierarchies simultaneously. For instance, nodes near the root of the tree under one relation may be leaf nodes under another. Besides,  as mentioned in~\cite{kolyvakis2019hyperkg}, Kiroukov \etal have shown that this kind of complex networks naturally emerge in the hyperbolic space, which provides a new way to model the data~\cite{krioukov2010hyperbolic}. 

Most of the previous works~\cite{bordes2013translating,yang2015embedding} model in the Euclidean space, relying on the inner product or Euclidean distance as a similarity measure, which can be categorised as translational models and bilinear models, respectively. Work~\cite{suzuki2018riemannian} proposed to embed the entities in a Riemannian manifold, where each relation is modeled as a move to a point and they also defined specific novel distance dissimilarities for the relations. However, as pointed out by~\cite{balazevic2019multi}, this model defined in the hyperbolic space does not outperform Euclidean models. The challenge for representing multi-relational data lies in the difficulty to represent entities, shared across relations, such that different hierarchies are formed under different relations. To deal with these issues, work~\cite{balazevic2019multi} presented a new bilinear model called MuRP to embed hierarchical multi-relational data in the Poincar\'e ball model of hyperbolic space. MuRP defines a basis score function for multi-relational graph embedding and generalizes it to the hyperbolic space. Experiments show that MuRP can get superior performances on the link prediction task. Besides, it requires far fewer dimensions than Euclidean embeddings to achieve comparable performance. Work~\cite{chami2020low} points out that MuRP is not able to encode some logical properties of relationships. Therefore, the authors leveraged the hyperbolic isometries to simultaneously exhibit logical patterns and hierarchies, and achieved the current best performance.

On the contrary, work~\cite{kolyvakis2019hyperkg} proposed a new translational model in Poincar\'e model, where both the entities and the relations are embedded. Compared to the translational models in Euclidean space, this method almost doubles the performance in terms of the mean reciprocal rank (MRR) metric. However, as mentioned by the authors, the HyperKG excludes from the comparison with many recent works that explores advanced techniques, thus this method is not comparable to the state-of-the-art methods, as listed in results in~\cite{balazevic2019multi}.

\subsection{Hyperbolic for Computer Vision}

\begin{figure}[t]
\centering
\includegraphics[width=0.4\textwidth]{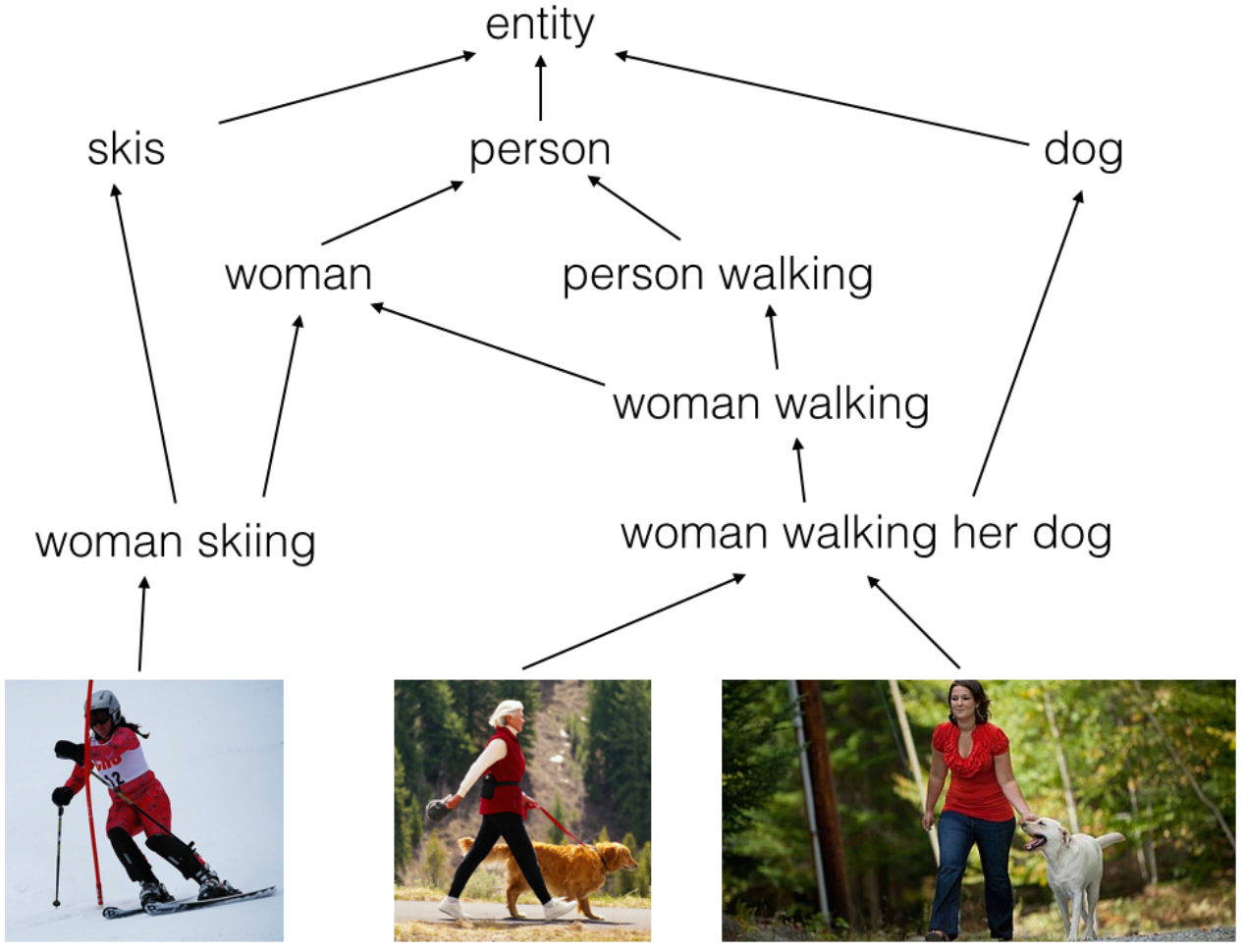}
\caption {\small{Example of hierarchy in computer vision~\cite{vendrov2016order}}. In combination with NLP, hierarchies can also be constructed in computer vision tasks. }
\label{Fig:visionhyperbolic}
\end{figure}

As mentioned above, hyperbolic space fits very well for NLP tasks due to their natural ability to embed the hierarchies. Compared to NLP applications, there are limited works for small-scaled computer vision tasks in hyperbolic space.  However, with these few available studies~\cite{khrulkov2020hyperbolic,skopek2020mixed,weber2020robust,chami2020trees,mathieu2019continuous,grattarola2019adversarial,ovinnikov2019poincar,peng2020mix}, we can still witness the potential of modeling computer vision tasks in the non-Euclidean space. The passion of doing this is inspired from the observation that similar hierarchical relations between images are also common in computer vision tasks~\cite{khrulkov2020hyperbolic}. For instance, image retrieval tasks naturally can be modeled by tree-structure since the overview photograph is related to many images that correspond to the close-ups of different distinct details. Besides,  hierarchies investigated in NLP can be also transcended to the visual domain, like the knowledge graph, as illustrated in Fig.~\ref{Fig:visionhyperbolic}.

Work~\cite{khrulkov2020hyperbolic} is one of the pioneering methods to model images in the hyperbolic space. As the author mentioned, the penultimate layer of a deep neural network is utilized to learn higher level semantic representations for image embedding. Then the succeeding latter operations can be treated as implying a certain type of geometry of the embedding spaces, and pairs of classes are separated by Euclidean hyperplanes. To prove it is reasonable to utilize the hyperbolic space, they turned to the $\delta$-Hyperbolicity, which is used to measure the basic common property of “negatively curved” spaces and where $\delta$ is defined as the minimal value such that the Gromov product holds for all points (features). The smaller the $\delta$ is, the higher the hyperbolicity of a dataset is. By using this, the author specified how close are the datasets CIFAR10, CIFAR100~\cite{krizhevsky2009learning}, CUB~\cite{wah2011caltech} and MiniImageNet to a hyperbolic space. They proved that the feature embeddings from current famous architectures like ResNet~\cite{he2016deep}, VGG19~\cite{simonyan2014very}, and InceptionV3~\cite{szegedy2015going} are with a small $\delta$ thus processing some kind of hyperbolicity. Therefore, learning image embedding in hyperbolic space can be benefited. 

Based on these observations, the authors constructed the analogues of layers in hyperbolic spaces. They evaluated their models in computer vision tasks, including Person re-identification~\cite{zheng2016person}, and few–shot classification~\cite{snell2017prototypical}, and results proved its superiority.   

Another interesting observation from work~\cite{khrulkov2020hyperbolic} is the measurement of uncertainty. The author found that the distance to the center in Poincar\'e ball indicates a model's uncertainty. As pointed out by another work~\cite{verma2019manifold}, one common problem of current neural networks is that they often provide incorrect but very confident predictions when evaluated on slightly different test examples. Therefore, this finding may also improve the robustness of current neural networks via either measuring the uncertainty or smoothing the decision boundary. 

Work~\cite{peng2020mix} introduced a Poincar\'e ST-GCN for the skeleton-based action recognition task. Since the input sequence (skeletons) can be modelled by the graph, thus compared to a normal computer vision task with images/RGB video input,  it is straightforward to tend to hyperbolic space. But one thing to notice is that the author modelled the skeleton frame-by-frame, instead of the whole sequence as a point in the hyperbolic space.

Another study in computer vision is trying to generalize generative models like VAE to the hyperbolic space and deal with image reconstruction or generation tasks.  Work~\cite{ovinnikov2019poincar} provided a Wasserstein Autoencoder on the Poincar\'e model and apply it to the task of generating binarized MNIST digits in order to get an intuition for the properties of the latent hyperbolic geometry. However, they did not provide a better reconstruction results when compared to the Euclidean counterpart. As mentioned by the authors, both the models meet with a dimension mismatch problem~\cite{rubenstein2018latent} such that reconstructed samples present a deteriorating quality as the dimensionality increases despite the lower reconstruction error. But, this problem is worse for the Poincar\'e model, as shown in the illustrations of work~\cite{ovinnikov2019poincar}. Compared to  work~\cite{mathieu2019continuous}, the authors derived a closed-form solution of the ELBO with two different kinds of normal distributions in the hyperbolic space. Their VAE model outperforms its Euclidean counterpart, especially for low latent dimensions. However, as the latent dimension increases, the embeddings quality decreases, hence the gain from the hyperbolic geometry is reduced, just as also observed by work~\cite{nickel2017poincare}. The same situation is also found in the work~\cite{skopek2020mixed}, where a mixed geometry space is introduced. On the MINIST reconstruction task, they displayed clear advantage when setting the latent dimension to six. However, this advantage to Euclidean space immediately disappears when they double the dimension of latent space. This can be also caused by the property of the datasets. More studies are needed to answer whether hyperbolic space has its advantages to address computer vision tasks. Moreover, it needs to extend to more complicated settings in larger-scale cases.

\section{Discussion and Open Problems}\label{Sec7}

\subsection{Which hyperbolic models?}
Though both Lorentz model and Poincar\'e model are commonly used in generalizing neural networks, we can find majority of the research in the literature is based on the latter one. However, some current research found that the Lorentz model has shown to have better numerical stability properties~\cite{nickel2018learning}, due to its large variance when close to the boundary. Recently, on the top of the Lorentz model with tiling method, Yu \etal~ provided a more stable model~\cite{yu2019numerically}. However, Lorentz model is un-bounded from the definition, which is not friendly to modern neural networks. Therefore, more studies are needed to ensure the right choice of hyperbolic model. Currently, we also find a trend to construct neural networks utilizing mixed geometries, as described in Sec.~\ref{sec:mixNN}. For the Neural Networks with mixed geometries, there is a great potential especially for the real-world data possessing with very complicated structure or relationships. However, it is still not explored to a full extent. One major issue is how to construct the product manifold and at the same time determine its signature. Besides, current optimizations require a costly grid search to tune hyperparameters. Optimization is also a big problem of the neural network with mixed geometries.

\subsection{Hyperbolic Neural Networks for Computer Vision}

Computer vision is one core topic in machine learning and it is also one of our focuses. As mentioned before, hyperbolic neural networks are mainly constructed for NLP, graph and other tree-structured datasets. However, there is still limited research for generalizing convolutional networks to hyperbolic space for image/video based tasks. One possible reason is that the conventional CNNs have already done a great job for the conventional computer vision tasks. And many hyperbolic counterparts do not share the superiority, especially when the embedding dimension is high. Another reason may lie in the fact that the tree-like structure is not significant as it for data like graph or symbolic natural language. Therefore, one potential application for image/video database is to explore the top-stream tasks like relationship inference, instead of solving the down-stream tasks like image classification. Another more attractive direction can be learning a hierarchical feature representation, instead of proving limited constraints to the feature learning processing. This also satisfies the concepts of cognitive science. In this way, it definitely can be conducted in the hyperbolic space.

\subsection{Consistency and Stability}
Another interesting observation is that the use of hyperbolic embeddings improves the performance for some problems and datasets, while not helping others~\cite{ganea2018hyperbolic}. For many cases, the Euclidean counterparts show superiority especially when embedding dimensions become larger. It is not clear whether the hyperbolic model can only provide a more compact model or it can provide a more efficient model with significant performance improvement. At the same time, a better understanding of when and why the use of hyperbolic geometry is justified is critical for this moment and therefore is very needed. Another problem is that different methods (even for the same backbone) test their methods using different tasks, or the same task with different protocols. It is hard to make the comparison. Thus, a standard benchmark for evaluating different hyperbolic approaches are also needed.

\subsection{Advanced Hyperbolic Networks}
One potential direction can be the combination of Riemannian neural network with advanced deep learning technology. For instance, exploring new Riemannian neural architectures with advanced automatic machine learning methods, like NAS~\cite{elsken2019neural}.  Work~\cite{peng2020mix} provided to search the best projection dimension in the Poincar\'e model, utilizing the NAS method.  However, there is much room to improve by automatically designing the neural modules, instead of only searching for optimal projection dimensions.
Another important research topic can be the generalization of more sophisticated Euclidean optimization algorithms. In many cases, as mentioned in~\cite{ganea2018hyperbolic}, fully Euclidean baseline models might have an advantage over hyperbolic baselines. One possible reason is that Euclidean space is equipped with much more advanced optimization tools. Once the hyperbolic neural networks are also equipped with advanced machine learning tools, we can expect more advanced and powerful hyperbolic networks.


%



\ifCLASSOPTIONcompsoc
  \section*{Acknowledgments}
\else
  \section*{Acknowledgment}
\fi

The authors would like to thank Octavian-Eugen Ganea, from MIT  for the useful discussion. We also want to thank Emile Mathieu, from University of Oxford, for the explanation regarding the gyroplane layer in their Poincar\'e Variational Auto-Encoder.

This work is supported by the Academy of Finland for ICT 2023 project (grant 328115) and project MiGA (grant 316765) and Infotech Oulu. 

\ifCLASSOPTIONcaptionsoff
  \newpage
\fi



%
\ifCLASSOPTIONcaptionsoff
  \newpage
\fi

\bibliographystyle{IEEEtran}
\bibliography{citation}

%



\newpage

\begin{IEEEbiographynophoto}{Wei Peng}
is currently a Ph.D. candidate with the Center for Machine Vision and Signal Analysis, University of Oulu, Oulu, Finland.  He received the M.S. degree in computer science from the Xiamen University, Xiamen, China, in 2016. His articles have published in mainstream conferences and journals, such as AAAI, ICCV, ACM Multimedia, Transactions on Image Processing. His current research interests include machine learning, affective computing, medical imaging, and human action analysis.
\end{IEEEbiographynophoto}

\begin{IEEEbiographynophoto}{Tuomas Varanka}
is currently a Ph.D. candidate with the Center for Machine Vision and Signal Analysis, University of Oulu, Oulu, Finland. He received his B.S. and M.S. degree in computer science and engineering from the University of Oulu in 2019 and 2020, respectively. His work has focused on micro-expression recognition. His current research interests include machine learning, and affective computing.
\end{IEEEbiographynophoto}

\begin{IEEEbiographynophoto}{Abdelrahman Mostafa}
received the M.S. degree in Artificial Intelligence, University of Oulu, Finland. He is currently a researcher and PhD student, University of Oulu, Finland. His research interests are Machine Learning, Deep Learning and Computer Vision.
\end{IEEEbiographynophoto}

\begin{IEEEbiographynophoto}{Henglin Shi}
received the M.S. degree in Artificial Intelligence, University of Oulu, Finland. He is currently a researcher and PhD student, University of Oulu, Finland. His articles have published in mainstream conferences and journals, such as BMVC, TMM and TIP. His research interests are Machine Learning, Deep Learning and Computer Vision.
\end{IEEEbiographynophoto}

\begin{IEEEbiographynophoto}{Guoying Zhao}
received the Ph.D. degree (2005) in computer science from the Chinese Academy of Sciences, Beijing, China. She is currently a full professor with the Center for Machine Vision and Signal Analysis, University of Oulu, Finland. Her work has focused on affective computing and machine learning. She is IEEE senior member, IAPR fellow and ELLIS member, and associate editor for Pattern Recognition, IEEE Transactions on Circuits and Systems for Video Technology, and Image and Vision Computing Journals.
\end{IEEEbiographynophoto}






\end{document}